\documentclass[preprint,12pt,authoryear]{elsarticle}
\usepackage{amssymb}
\usepackage{hyperref} 
\usepackage{lineno}
\usepackage{soul}
\usepackage{setspace}
\usepackage{subfig}
\usepackage{bm}
\usepackage{amssymb}
\usepackage{amsmath}
\usepackage{upgreek}
\usepackage{bigints}
\usepackage{algorithm2e}
\usepackage{algpseudocode}
\usepackage{url}
\usepackage{hyphenat}
\RestyleAlgo{ruled}
\usepackage{float}
\usepackage[left=2.1cm, right=2.1cm, top=3.0cm, bottom=3.0cm]{geometry}

\journal{Geoenergy Science and Engineering}
\begin{document}
\begin{frontmatter}

\title{Accelerated training of deep learning surrogate models for surface displacement and flow, with application to MCMC-based history matching of CO$_2$ storage operations} 

\author[inst1]{Yifu Han}
\affiliation[inst1]{organization={Department of Energy Science and Engineering}, 
            addressline={Stanford University}, 
            city={Stanford},
            state={CA},
            postcode={94305}, 
            country={USA}}

\author[inst2]{Fran\c cois P. Hamon}
\affiliation[inst2]{organization={TotalEnergies, E\&P Research and Technology},
            addressline={1201 Louisiana Street},
            city={Houston},
            state={TX},
            postcode={77002},
            country={USA}}

\author[inst1]{Louis J. Durlofsky}

\begin{abstract}

Deep learning surrogate modeling shows great promise for subsurface flow applications, but the training demands can be substantial. Here we introduce a new surrogate modeling framework to predict CO$_2$ saturation, pressure and surface displacement for use in the history matching of carbon storage operations. Rather than train using a large number of expensive coupled flow-geomechanics simulation runs (required to predict surface displacement), training here involves a large number of inexpensive flow-only simulations combined with a much smaller number of coupled runs. The flow-only runs use an effective rock compressibility, which is shown to provide accurate predictions for saturation and pressure for our system. A recurrent residual U-Net architecture is applied for the saturation and pressure surrogate models, while a new residual U-Net model is introduced to predict surface displacement. The surface displacement surrogate accepts, as inputs, geomodel quantities along with saturation and pressure surrogate predictions. Median relative error for a diverse test set is less than 4\% for all variables. The surrogate models are incorporated into a hierarchical Markov chain Monte Carlo history matching workflow (this workflow would be infeasible with high-fidelity simulation). Surrogate error is included using a new treatment involving the full model error covariance matrix. A high degree of prior uncertainty, with geomodels characterized by uncertain geological scenario parameters (metaparameters) and associated realizations, is considered. History matching results for a synthetic true model are generated using in-situ monitoring-well data only, surface displacement data only, and both data types. The enhanced uncertainty reduction achieved with both data types is quantified. Posterior saturation and surface displacement fields are shown to correspond well with the true solution. The impact of properly treating surrogate model error, and the performance of the workflow for another, more challenging true model, are presented in Supplemental Information.

\end{abstract}


\end{frontmatter}

\section{Introduction}
\label{Introduction}
Large-scale CO$_2$ injection is likely to induce geomechanical responses, including some amount of surface displacement. Surface uplift can be measured by satellite, and the resulting dynamic data may enable an improved understanding of the geological carbon storage operation. Simulations that involve coupled flow and geomechanics are required, however, if displacement data are to be incorporated into a formal history matching procedure. This could represent a challenge because the computational cost associated with coupled simulations can be large. The use of a fast and accurate surrogate model that can be constructed efficiently would thus be highly useful in this setting.

In this study, we introduce surrogate modeling procedures to provide predictions of flow and surface displacement quantities for the history matching of large-scale storage operations. A key advantage of our surrogate model is that relatively few coupled flow-geomechanics simulations are required for training. A substantial number of flow-only simulations are needed, though these are much less expensive than the coupled runs, and we show how they can be used with a small set of coupled runs to provide accurate surface displacements. The geomodels used for training are characterized by uncertain geological scenario parameters, referred to here as metaparameters. The surrogate models are used in a hierarchical Markov chain Monte Carlo (MCMC) history matching procedure~\citep{han2023surrogate}, with model error included in the framework. The impact of different data types (e.g., surface and subsurface measurements) on posterior results is evaluated.

Geomechanical effects in geological carbon storage problems have been investigated by a number of researchers. \citet{vilarrasa2016geomechanical} evaluated caprock stability during injection using stress data from fully coupled simulations. \citet{jahandideh2021inference} performed coupled simulations to predict rock failure and microseismicity from stress and strain data. \citet{li2016coupled} assessed the impact of the storage-aquifer Biot coefficient on surface displacement predictions for a geomodel based on In Salah. The predicted surface vertical displacement generally agreed with satellite-based displacement measurements. \citet{zheng2021geologic} used coupled flow-geomechanics simulations for multiobjective optimization. They developed an optimization framework to determine injection well locations that maximize the amount of injected CO$_2$ while minimizing vertical displacement and plastic strain. \citet{rahman2022effect} showed that the spatial variation of overburden rock properties can have a significant effect on vertical displacement. \citet{sun2023geomechanical} developed a workflow to assess injection-induced geomechanical effects. They evaluated the impact of uncertain rock physical properties and initial stress on geomechanical response. \citet{alsayah2023coupled} considered systems with shale inter-layers in the storage aquifer, and assessed how they affect CO$_2$ plume migration, vertical displacement and stress response.

A number of surrogate models have been developed to predict flow and geomechanical responses in coupled systems. \citet{yoon2021modeling} developed a U-Net-based surrogate model to provide pressure, saturation and vertical displacement fields in 3D storage aquifers. Promising results were achieved for idealized models.
\citet{tang2022deep} extended their earlier recurrent R-U-Net surrogate model~\citep{tang2020deep,tang2021deep} to treat coupled 3D systems. Their model provided (3D) pressure and saturation fields in the storage aquifer along with vertical displacement at the Earth's surface. They considered only a single geological scenario, with multi-Gaussian realizations of log-permeability. Similarly, \citet{2022deep} used an R-U-Net-based surrogate model to predict 2D surface displacement and spatially averaged pressure fields. Surrogate model inputs included 3D bimodal log-permeability and porosity fields, and time. \citet{zheng2024deep} considered the nonlinear plastic behavior of the geomechanical response. They developed a workflow to sequentially train two U-Net surrogate models with permeability and injection well locations as input. The first model provides pressure, saturation and vertical displacement fields, while the second model predicts plastic strain fields using results from the first model as input.  

Deep-learning-based surrogate models have been used in a number of workflows focused on history matching for geological carbon storage. We first consider frameworks that did not include geomechanical effects. \citet{seabra2024ai} evaluated the performance of two surrogate models, Fourier neural operators (FNO) and transformer U-Net (T-UNet), for a 2D channelized storage aquifer with uncertain channel orientation and sinuosity. The surrogate models were used in an ensemble smoother with multiple data assimilation (ESMDA) workflow to reduce uncertainty in the permeability field. \citet{he2024deep} developed a surrogate model to predict pressure in a monitoring well, which was then used with MCMC-based data assimilation to locate CO$_2$ leaks. In recent work, we extended the recurrent R-U-Net surrogate model and incorporated it in a hierarchical MCMC-based history matching workflow~\citep{han2023surrogate}. Uncertainty in the geological metaparameters was considered in that study. Surrogate models have also been applied for history matching in systems with coupled flow and geomechanics. \citet{tang2022deep} used vertical surface displacement as observation data in a rejection sampling-based history matching workflow. A single geological scenario was considered in that study. Similarly, \citet{2022deep} applied their surface displacement surrogate model in an ensemble-based history matching procedure to reduce uncertainty in bimodal porosity and permeability fields.

Our first goal in this study is the development of a new surrogate modeling framework that does not require an excessive number of coupled flow-geomechanics runs for training. We accomplish this by introducing a new network architecture for the surface displacement variables, which differs from that applied for the flow variables (pressure and saturation in the storage aquifer). This enables us to use a factor of 10 fewer coupled runs than flow-only runs in the training step, which results in substantial computational savings. We also demonstrate the accuracy of an effective rock compressibility treatment in the flow-only simulations.

Our second key goal is to use our hierarchical MCMC history matching procedure with the surrogate models to assess the impact of different data types on posterior results. The prior geomodels, used for network training, are characterized by several uncertain metaparameters, including the mean and standard deviation of log-permeability, permeability anisotropy ratio, constants relating porosity to log-permeability, and Young's moduli in the storage aquifer and overburden. It is important to emphasize that each set of metaparameters corresponds to an infinite number of geomodel realizations, so a high degree of prior uncertainty is considered in our modeling. We compare posterior predictions for the metaparameters for three different approaches -- using only in-situ monitoring-well data, only surface displacement data, and both data types together. A procedure for treating model error in the data assimilation framework is also introduced. Although deep learning surrogate models are now widely used for history matching, the model error associated with them does not appear to have been considered in previous studies.

This paper proceeds as follows. In Section~\ref{Geomodel}, we describe the basic setup and the geological scenarios considered in this work. The flow-only and fully coupled simulation runs, all performed using GEOS~\citep{bui2021multigrid}, are discussed in Section~\ref{Comparison}. In Section~\ref{Surrogate Model}, the surrogate modeling procedure, including the detailed network architectures and training strategies, are presented. Surrogate model predictions for pressure, saturation and surface displacement are compared to reference GEOS results in Section~\ref{Surrogate-performance}. In Section~\ref{History-matching}, we present history matching results using the hierarchical MCMC method for a synthetic true model. The impact of different data types on posterior results is assessed. Conclusions and ideas for future work are provided in Section~\ref{Conclusions}. In Supplementary Information (SI), available online, we provide history matching results with model error neglected, as well as results for a true model characterized by a correlation structure that differs from the prior.

\section{Geomodel and Simulation Setup}
\label{Geomodel}
In this section, we first discuss the metaparameters that characterize the geological scenarios and the procedures used to construct geomodel realizations. The basic simulation setup is then described.

\subsection{Geomodel and metaparameters}
\label{Geomodels}

\begin{figure}[!ht]
\centering  
\includegraphics[width=15cm]{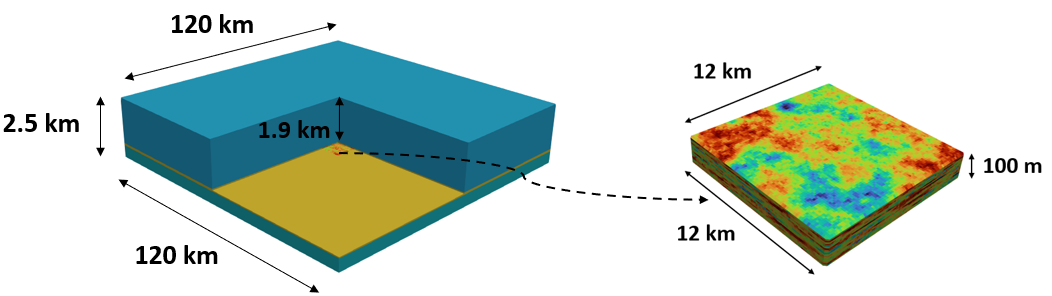}
\caption{Model domain used for both the flow-only and coupled flow-geomechanics simulations. The full domain includes the storage aquifer, surrounding region, overburden and underburden. A realization of the storage aquifer is shown on the right.}
\label{Model_Setup}
\end{figure}

The geomodel considered in this study is based on models of the Mt.~Simon sandstone formation, under evaluation for the Decatur storage project in the Illinois Basin~\citep{okwen2022storage}. The full domain, shown in Fig.~\ref{Model_Setup}, includes the central storage aquifer, a large lateral surrounding region, and overburden and underburden rock. The overall domain is of size 120~km $\times$ 120~km $\times$ 2.5~km, and is represented on a simulation grid of 100 $\times$ 100 $\times$ 30 cells (total of 300,000 cells). The depth to the top of the storage aquifer is 1900~m, consistent with the Mt.~Simon  formation~\citep{okwen2022storage}. The central storage aquifer is characterized by multi-Gaussian log-permeability and porosity fields. It is of size 12~km $\times$ 12~km $\times$ 100~m, and is represented on a uniform grid of 80 $\times$ 80 $\times$ 20 cells (total of 128,000 cells). The cell size in the surrounding region increases with distance from the storage aquifer.

In this work, we construct geomodel realizations based on geological scenarios characterized by seven metaparameters. An infinite number of realizations of the full domain can be generated for each set of metaparameters, so a high degree of geological variability and prior uncertainty is considered in our workflow. The metaparameters, denoted $\boldsymbol{\uptheta}_{\mathrm{meta}} \in \mathbb{R}^{7}$, are as follows

\begin{equation} \label{metaparameters}
\boldsymbol{\uptheta}_{\mathrm{meta}} = [\mu_{\log k}, \sigma_{\log k}, a_r, d, e, E_s, E_o].
\end{equation}
\noindent In Eq.~\ref{metaparameters}, $\mu_{\log k}$ and $\sigma_{\log k}$ are the mean and standard deviation of the multi-Gaussian log-permeability field in the storage aquifer, $a_r$ is the permeability anisotropy ratio, $d$ and $e$ are coefficients relating porosity to log-permeability, and $E_s$ and $E_o$ represent the constant-in-space Young's moduli of the storage aquifer and overburden/underburden rock, respectively. 

Here, as in \citet{han2023surrogate}, we generate realizations of the log-permeability field for the storage aquifer by rescaling standard normal fields of fixed correlation structure. The rescaling acts to shift the mean and standard deviation to the values prescribed by metaparameters $\mu_{\log k}$ and $\sigma_{\log k}$. The horizontal ($x$ and $y$) correlation lengths are set to 4800~m (32~cells), and the vertical correlation length ($z$) is set to 40~m (8~cells). These (large) correlation lengths are consistent with those used in previous models of the Mt.~Simon formation \citep{crain2023integrated}.

Realizations of the standard normal fields are generated using principal component analysis (PCA). The PCA basis matrix is constructed from a set of 1000 unconditioned realizations generated using the geological modeling software SGeMS~\citep{remy2009applied}. An exponential variogram model, with correlation lengths given above, is used. Once the PCA model is constructed, new realizations can be generated through application of

\begin{equation} \label{pca}
\textbf{y}^{\mathrm{pca}} = \Phi \bm{\xi} + \bar{\textbf{y}}, 
\end{equation} 
\noindent where $\Phi \in \mathbb{R}^{n_s \times n_d}$ is the basis matrix truncated to $n_d$ columns (we use $n_d=800$ in this study), $\bar{\textbf{y}} \in \mathbb{R}^{n_s}$ is the mean of the SGeMS realizations used to generate the basis matrix $\Phi$, and $\bm{\xi} \in \mathbb{R}^{n_d}$ is the low-dimensional standard-normal PCA latent variable. For details on the PCA procedure and a discussion of the time savings achieved through use of this approach, please see \citet{han2023surrogate}.

Given $\textbf{y}^{\mathrm{pca}} \in \mathbb{R}^{n_s}$, where $n_s$ = 128,000 is the number of cells in the storage aquifer, the full geomodel is described as follows. For cell $i$ ($i = 1, 2, \dots, n_s$) in the storage aquifer, the permeability $(k_s)_i$ is given by $(k_s)_i = \exp \left( \sigma_{\log k} \cdot (y^{\mathrm{pca}})_i + \mu_{\log k} \right)$. We set $(k_x)_i=(k_y)_i=(k_s)_i$ and $(k_z)_i=a_r \cdot (k_s)_i$, where $(k_x)_i$, $(k_y)_i$ and $(k_z)_i$ denote directional permeabilities for cell $i$. Porosity $(\phi_s)_i$ is assigned through application of $(\phi_s)_i = d \cdot \log_e(k_s)_i + e$. Young's modulus in the storage aquifer ($E_s$) is taken to be constant in space, though this value differs between realizations. Poisson’s ratio in the storage aquifer is set to 0.165 in all realizations, consistent with the value reported for the Mt.~Simon formation by \citet{bauer2016overview}. The ranges for all metaparameters, along with the values for key fixed parameters, are given in Table~\ref{rock physics}.

The properties of the surrounding region, overburden and underburden are taken to be constant in space. The porosity and log-permeability of the surrounding region are set to the mean porosity and mean log-permeability of the storage aquifer. Young's modulus and Poisson’s ratio in the surrounding region are set to be the same as in the storage aquifer. 

The porosity and permeability in the overburden are set to 0.08 and 0.001~md, and in the underburden they are specified as 0.09 and 2.3~md. These values derive from \citet{bauer2016overview}. Young's modulus ($E_o$) is again taken to be constant in space but variable between realizations, with the same value used for the overburden and underburden. The allowable values for $E_o$, shown in Table~\ref{rock physics}, are larger than those for $E_s$, consistent with the lower porosity in the overburden and underburden rock \citep{babarinde2021workflow}. Poisson’s ratio for the overburden is set to 0.27. This corresponds to the value reported for the Eau Claire formation, which represents the overburden for the Mt.~Simon formation~\citep{bauer2016overview}. Poisson’s ratio for the underburden is also set to 0.27.

\begin{table}
\begin{center}
\footnotesize
\caption{Parameters used in the coupled flow and geomechanics simulations}
\label{rock physics}
\renewcommand{\arraystretch}{1.3} 
\begin{tabular}{ c c } 
\hline
\textbf{Overburden parameters} & \textbf{Values} \\ 
\hline
Thickness & 1905 m \\
Permeability & 0.001 md \\ 
Porosity & 0.08 \\ 
Young's moduli ($E_o$) & $E_o$ $\sim$ $U$(25, 40) GPa \\
Poisson's ratio & 0.27 \\
\hline
\textbf{Storage aquifer parameters} & \textbf{Values or ranges} \\ 
\hline
Thickness & 100 m \\
Horizontal correlation lengths ($l_h$) & 32~cells (4800~m) \\
Vertical correlation length ($l_v$) & 8~cells (40~m) \\
Mean of log-permeability ($\mu_{\log k}$)  & $\mu_{\log k} \sim$ $U$(2, 5): (7.4, 148.4) md \\ 
Standard deviation of log-permeability ($\sigma_{\log k}$) & $\sigma_{\log k} \sim$ $U$(1, 2.5) \\ 
Parameter $d$ in $\log k$ -- $\phi$ correlation & $d \sim U(0.02, 0.04)$ \\ 
Parameter $e$ in $\log k$ -- $\phi$ correlation & $e \sim U(0.06, 0.08)$ \\ 
Permeability anisotropy ratio ($a_r$)  & $\log_{10}(a_r)$ $\sim$ $U$(-2, 0) \\
Young's moduli ($E_{s}$) & $E_{s}$ $\sim$ $U$(5, 20) GPa \\
Poisson's ratio & 0.165 \\
\hline
\textbf{Surrounding region parameters} & \textbf{Values} \\ 
\hline
Thickness & 100 m \\ 
Permeability & $\exp(\mu_{\log k})$ md \\ 
Porosity & $\bar{\phi}_s$ \\ 
Young's modulus & $E_{s}$ $\sim$ $U$(5, 20) GPa \\
Poisson's ratio & 0.165 \\
\hline
\textbf{Underburden parameters} & \textbf{Values} \\ 
\hline
Thickness & 500 m \\
Permeability & 2.3 md \\ 
Porosity & 0.09 \\ 
Young's modulus & $E_o$ $\sim$ $U$(25, 40) GPa \\
Poisson's ratio & 0.27 \\
\hline
\textbf{Relative permeability and capillary pressure} & \textbf{Values} \\ 
\hline
Irreducible water saturation ($\emph{S$_{wi}$}$) & 0.22 \\
Residual CO$_2$ saturation ($\emph{S$_{gr}$}$) & 0 \\
Water exponent for Corey model ($\emph{n$_{w}$}$) & 9 \\
CO$_2$ exponent for Corey model ($\emph{n$_{g}$}$) & 4 \\
Relative permeability of CO$_2$ at $\emph{S$_{wi}$}$ ($\emph{k$_{rg}$}$($\emph{S$_{wi}$}$)) & 0.95 \\
Capillary pressure exponent ($\lambda$) & 0.55 \\ 
\hline
\end{tabular}
\end{center}
\end{table}

Geomodel realizations for the full domain are denoted $\textbf{m}_f \in$ $\mathbb{R}^{n_c \times n_p}$, where $n_c = 300,000$ is the number of cells in the full domain and $n_p=4$ is the number of properties in each cell that vary between realizations. These properties correspond to horizontal permeability, vertical permeability, porosity, and Young's modulus. Realizations are constructed from the metaparameters ${\boldsymbol{\uptheta}}_{\mathrm{meta}}$ and the PCA latent variables $\bm{\xi}$ appearing in Eq.~\ref{pca}. This overall representation can be expressed as $\textbf{m}_f = \textbf{m}_f\left({\boldsymbol{\uptheta}}_{\mathrm{meta}}, \textbf{y}^{\mathrm{pca}}(\bm{\xi})\right)$.

During the construction of random geomodel realizations, which are used for training and testing of the surrogate models and for prior models in history matching, we sample $\mu_{\log k}$, $\sigma_{\log k}$, $d$, $e$, $E_s$ and $E_o$ from the uniform distributions given in Table~\ref{rock physics}. The permeability anisotropy ratio of the storage aquifer ($a_r$) is sampled from a log-uniform prior distribution. During history matching, the goal is to identify posterior samples, which then enable us to estimate the posterior distribution. The allowable posterior parameter ranges are the same as those defined in Table~\ref{rock physics}.

\subsection{General simulation setup}
\label{Simulation Setup}

\begin{figure}[!ht]
\centering   
\subfloat[Injection and observation well locations]{\label{wells}\includegraphics[width = 78mm]{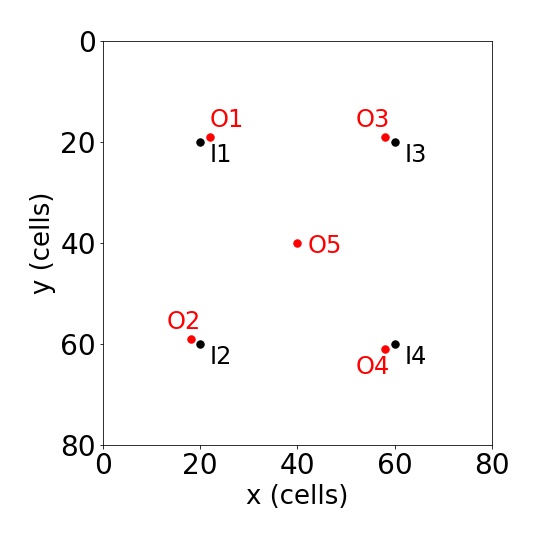}}
\hspace{5mm}
\subfloat[Surface displacement observation locations]{\label{surface}\includegraphics[width = 78mm]{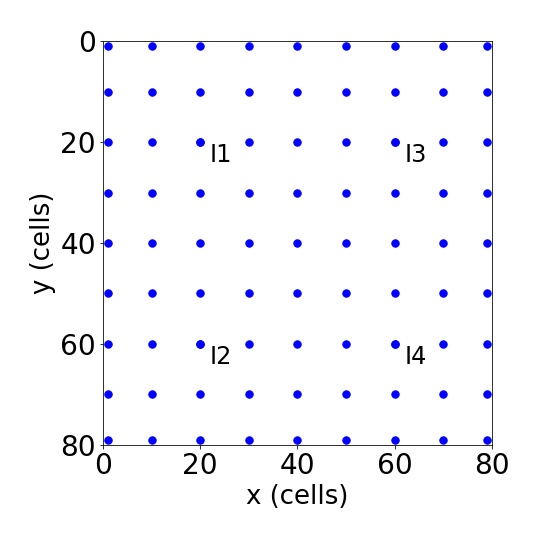}}
\caption{Locations of injection wells (I1--I4) and observation wells (O1--O5) in the storage aquifer are shown in (a). Locations of the surface vertical displacement measurements (directly above the storage aquifer) shown in (b).}
\label{observations}
\end{figure}

In this study, we consider a constant-rate injection scenario involving four fully penetrating vertical wells. Each well injects 1~Mt CO$_2$/year (total injection of 4~Mt/year) over a 30-year period. This is consistent with our previous setup for  flow-only problems~\citep{han2023surrogate}, and represents more injection than is initially planned for the Decatur project. This volume of injection is, however, in alignment with potential future scenarios. The locations of the four vertical injection wells (denoted I1--I4), along with the locations of five vertical observation wells (O1--O5), are indicated in Fig.~\ref{wells}. The 81 surface locations at which vertical displacement is measured are shown in Fig.~\ref{surface}. These locations are spaced every 1.5~km in the $x$ and $y$ directions.

Both the flow-only and the coupled flow-geomechanics simulations are performed using the GEOS simulator~\citep{bui2021multigrid}. Consistent with \citet{okwen2022storage}, the initial hydrostatic pressure and temperature in the middle layer (layer~10) of the storage aquifer are set to 20~MPa and 50.3~$^{\circ}$C. No-flow boundary conditions are applied at the boundaries of the full domain. Roller boundary conditions~\citep{chandrupatla2012introduction} are prescribed at the sides and bottom boundaries of the full domain in the coupled flow-geomechanics simulations. This means that displacement can occur in the plane of the boundary, but there is no displacement perpendicular to the boundary surface. Geochemical effects are not considered in our models.

\begin{figure}[!ht]
\centering
\subfloat[Relative permeability curves]{\label{relative_perm}\includegraphics[width = 79mm]{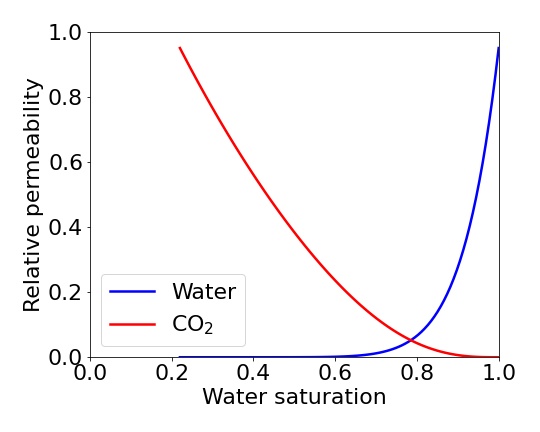}}
\hspace{5mm}
\subfloat[Capillary pressure curve]{\label{capillary}\includegraphics[width = 79mm]{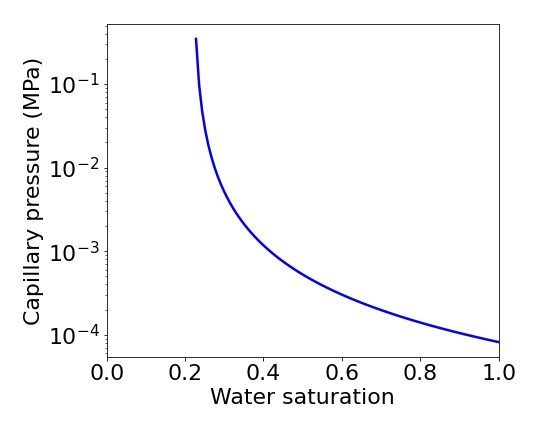}}
\caption{CO$_2$-brine relative permeability and capillary pressure curves used in all simulations. Capillary pressure curve in (b) is for a cell with porosity of 0.2 and permeability of 20~md.}
\label{Relative_Permeability}
\end{figure}

The relative permeability and capillary pressure curves used in all simulations are based on measured data from \citet{krevor2012}. The parameters for the relative permeability and capillary pressure functions used in this study are provided in Table~\ref{rock physics}. The capillary pressure curve is generated using the Brooks-Corey~\citep{saadatpoor2010new} model, with the capillary pressure for each cell computed from the Leverett J-function (which involves cell porosity and permeability). Relative permeability curves, along with a capillary pressure curve for a cell with a porosity of 0.2 and permeability of 20~md, are shown in Fig.~\ref{Relative_Permeability}.

\section{Flow-only and Coupled Simulation Comparisons}
\label{Comparison}
As noted earlier, GEOS is used to perform both the coupled flow-geomechanics and flow-only simulations. The latter are intended to provide storage aquifer pressure and saturation results that are in close agreement with those from coupled simulations. Because they do not model stress and displacement, the flow-only simulations are not able to provide predictions for surface displacement. In this section, we provide some background on the coupled simulations, describe the setup for the flow-only simulations, and compare results for storage aquifer pressure and saturation with the two formulations.

\subsection{Simulation formulations} \label{sec:formulations}

A displacement-pressure-composition formulation~\citep{huang2023fully} is applied for the coupled flow-geomechanics simulations. Two-phase CO$_2$-brine flow is considered, with the system assumed to be poroelastic and isothermal. The governing equations consist of a linear momentum balance equation for geomechanics, and mass balance equations for the two components for flow. The detailed equations are presented by \citet{t2022deformation} and \citet{tang2022deep}. At each simulation time step, the flow and geomechanical equations are solved using a fully implicit, fully coupled treatment. Displacement, pressure, composition, and saturation variables are updated until convergence is achieved. The geomechanical equations are solved using a finite element method, and the flow equations are solved using a finite volume procedure. In the linear momentum balance equation, changes in pressure impact the computation of total stress. In the mass balance equations for flow, changes in displacement impact  porosity ($\phi$) via

\begin{equation}
\Delta \phi = b \Delta \epsilon + \frac{(b - \phi^0)}{K_{g}} \Delta p. \label{eq:linearized_porosity_coupled}
\end{equation}
Here $\epsilon$ is volumetric strain, which is computed based on the symmetric gradient of the geomechanical displacement, $K_g$ is grain bulk modulus, $\phi^0$ is the initial porosity, and $b$ is the Biot coefficient, given by

\begin{equation} \label{Biot}
b = 1 - \frac{K}{K_g},
\end{equation}
where $K$ denotes the rock bulk modulus. This quantity is computed as $K = \frac{E}{3(1 - 2v)}$, where $E$ and $v$ are Young's modulus and Poisson’s ratio. In this study, the grain bulk modulus $K_g$ is set to 37~GPa in all cases. This value is corresponds to the bulk modulus of quartz~\citep{wang2015elasticity}, which is the dominant mineral in the Mt.~Simon sandstone~\citep{okwen2022storage}. 

Using the notation introduced by \citet{tang2022deep}, the high-fidelity fully coupled flow and geomechanics simulations on the full domain (Fig.~\ref{Model_Setup}, left) can be represented as

\begin{equation} \label{coupled forward}
\left[\textbf{p}_f, \enspace \textbf{S}_f, \enspace \textbf{d}_f\right] = f\left(\mathbf{m}_f\right),
\end{equation}
where $f$ denotes the high-fidelity fully coupled flow and geomechanics simulation, $\textbf{p}_f \in \mathbb{R}^{n_c \times n_{ts}}$, $\textbf{S}_f \in \mathbb{R}^{n_c \times n_{ts}}$ and $\textbf{d}_f \in \mathbb{R}^{3 \times n_{nd} \times n_{ts}}$ are the pressure, saturation, and displacement (in three coordinate directions) in the full domain at $n_{ts}$ simulation time steps, and $n_{nd} = 316,231$ is the number of finite element nodes in the full domain.

For the flow-only simulations, the compositional formulation involving isothermal two-phase CO$_2$-brine flow is applied~\citep{huang2023fully}. At each simulation time step, cell-centered pressure, composition and saturation are computed. As noted previously, relative permeability and capillary pressure curves are the same as in the coupled flow-geomechanics simulations. In the mass balance equations, changes in porosity are computed as a function of pressure and effective rock compressibility as follows:

\begin{equation}
\Delta \phi = c \Delta p. \label{eq:linearized_porosity_flow}
\end{equation}
where $c$ is the effective rock compressibility and $p$ is pore pressure. 

In many cases, a high degree of consistency between the coupled flow-geomechanics and flow-only solutions for saturation and pressure can be achieved through use of an effective rock compressibility that depends on the detailed geomechanical properties. This quantity is computed separately for each model domain (storage aquifer and lateral surrounding region, overburden, underburden) using the expression given by \citet{kim2010sequential}. Specifically, for the storage aquifer and surrounding region, the effective rock compressibility, denoted $c_s$, is given by

\begin{equation} \label{compressibility}
c_{s} = \frac{1 - 2v_{s}}{\bar{\phi}_s E_{s}} \left(b_s^2 \frac{1 + v_{s}}{1 -v_{s}} + 3(b_s - \bar{\phi}_s)(1 - b_s)\right),
\end{equation}
where $\bar{\phi}_s$ denotes the mean porosity for the storage aquifer, and $b_s$ and $v_s$ are the Biot coefficient and Poisson's ratio for the storage aquifer. Effective rock compressibilities for the overburden ($c_{o}$) and underburden ($c_{u}$) are computed analogously. The values for $c_{o}$ and $c_{u}$ differ slightly due to the different porosity values used for the two regions (see Table~\ref{rock physics}). With this treatment for effective rock compressibility,  flow-only simulations on the full domain can be represented as

\begin{equation} \label{flow forward}
\left[\widetilde{\textbf{p}}_f, \enspace \widetilde{\textbf{S}}_f\right] = \widetilde{f}\left(\mathbf{m}_f\right),
\end{equation}
where $\widetilde{f}$ represents the high-fidelity flow-only simulation, and $\widetilde{\textbf{p}}_f \in \mathbb{R}^{n_c \times n_{ts}}$ and $\widetilde{\textbf{S}}_f \in \mathbb{R}^{n_c \times n_{ts}}$ are pressure and saturation in the full domain at $n_{ts}$ simulation time steps.  

\subsection{Flow-only results using effective rock compressibility} \label{sec:p_S_comparison}

We now evaluate the accuracy of flow-only simulation results, for storage aquifer saturation and pressure, relative to reference coupled flow-geomechanics results. This assessment requires us to simulate, using both formulations, a large number of geomodel realizations of the full domain. Here we consider 500 randomly generated realizations, which are constructed by sampling the metaparameters from the distributions given in Table~\ref{rock physics}, and components of $\bm{\xi}$ (PCA latent variables) from ${\mathcal N}(0, 1)$. The same set of 500 realizations will be used as the test cases for evaluation of the surrogate models in Section~\ref{Surrogate-performance}.

The differences in CO$_2$ saturation and pressure in the storage aquifer between the two solutions are quantified over a set of time steps. These $n_t=10$ time steps -- which correspond to $t$ = 1, 2, 4, 6, 9, 12, 16, 20, 25 and 30~years --  are the same time steps as are used in the deep learning surrogate models. Saturation and pressure differences, denoted $\epsilon_S$ and $\epsilon_p$, are given by

\begin{equation} \label{difference_s}
\epsilon_S = \frac{1}{n_en_sn_t} \sum_{i=1}^{n_e} \sum_{j=1}^{n_s} \sum_{t=1}^{n_t} \frac{| (\widetilde{S}_s)_{i,j}^t - (S_s)_{i,j}^t |}{(S_s)_{i,j}^t + \epsilon}, 
\end{equation}
\begin{equation} \label{difference_p}
\epsilon_p = \frac{1}{n_en_sn_t} \sum_{i=1}^{n_e} \sum_{j=1}^{n_s} \sum_{t=1}^{n_t} \frac{| (\widetilde{p}_s)_{i,j}^t - (p_s)_{i,j}^t |}{(p_s)_{i,\mathrm{max}}^t - (p_s)_{i,\mathrm{min}}^t}. 
\end{equation}
Here $n_e$ = 500 is the number of geomodel realizations, $n_s$ = 128,000 is the total number of cells in the storage aquifer, $(\widetilde{S}_s)_{i,j}^t$ and $(\widetilde{p}_s)_{i,j}^t$ are CO$_2$ saturation and pressure predictions from the flow-only simulation, for geomodel realization $i$, storage aquifer cell $j$, and time step $t$. The quantities $(S_s)_{i,j}^t$ and $(p_s)_{i,j}^t$ are the corresponding predictions for the same geomodel realization from the coupled flow-geomechanics simulation. Here $(p_s)_{i,\mathrm{max}}^t$ and $(p_s)_{i,\mathrm{min}}^t$ are the maximum and minimum pressure (from the coupled flow-geomechanics runs) for all storage aquifer cells in realization $i$ at time step $t$. A constant ($\epsilon = 0.01$) is included in Eq.~\ref{difference_s} to prevent division by zero or a very small value.

The differences between the two solutions, computed using Eqs.~\ref{difference_s} and \ref{difference_p}, are $\epsilon_S=0.002$ (0.2\%) and $\epsilon_p=0.0007$ (0.07\%). These extremely small differences demonstrate the accuracy of the effective rock compressibility treatment for this case. Such close agreement would not be expected in more complicated settings involving, e.g., extensive fracturing and faulting, or plastic deformation. We reiterate that the effective rock compressibility models provide only saturation and pressure -- we still need fully coupled models to compute surface displacement.

The accuracy of the flow-only solutions is very beneficial in our case because it allows us to use these solutions for much of the training required by the surrogate models. With our current setup, a flow-only GEOS run requires about 8~minutes using 32 AMD EPYC-7543 CPU cores in parallel. Each fully coupled GEOS run, by contrast, requires about 120~minutes with the same 32 CPU cores. This increased computational time is partly due to the larger number of equations (momentum balance equations) and primary variables (displacements at the finite element nodes) in the coupled simulations, and partly because of the more advanced state of linear solver treatments in GEOS for flow-only runs. We expect the $15\times$ difference between the two approaches to decrease somewhat as more specialized strategies for the coupled problem become available in GEOS.

\section{Surrogate Modeling Procedure for Coupled Problems}
\label{Surrogate Model}
The surrogate models developed in this work provide very fast predictions for pressure and saturation in the storage aquifer and vertical displacement at the Earth's surface for new geomodel realizations characterized by the metaparameters described in Section~\ref{Geomodels}. These predictions are intended to accurately represent the corresponding solution from a fully coupled simulation of the input geomodel over the full domain. The well locations, injection rates and simulation time frame are the same in all runs. The architectures of the networks used for the prediction of storage aquifer pressure and saturation represent extensions of the architectures introduced by \citet{tang2022deep} and \citet{han2023surrogate}, as described below. Our approach for predicting surface displacement is different than in previous methods, and this new treatment enables us to use many fewer fully coupled runs for training.

\subsection{Surrogate model for pressure and saturation}
\label{sec:surr_press_sat}

We first describe the surrogate modeling procedure to approximate pressure and saturation in the storage aquifer. Two separate networks -- one for pressure and one for saturation -- are trained using solutions at 10 time steps ($t$ = 1, 2, 4, 6, 9, 12, 16, 20, 25 and 30~years) from flow-only simulations that utilize effective rock compressibilities (e.g., Eq.~\ref{compressibility}). As shown in Section~\ref{sec:p_S_comparison}, these solutions very closely approximate the pressure and saturation fields for the coupled problem. The surrogate model predictions for pressure and saturation can be expressed as 

\begin{equation} \label{flow surrogate}
\left[\hat{\textbf{p}}_{s}, \enspace \hat{\textbf{S}}_{s}\right] = \hat{f}\left(\mathbf{m}_f; \widetilde{\textbf{w}}\right),
\end{equation}
\noindent where $\hat{f}$ denotes the surrogate model, $\hat{\textbf{p}}_s \in \mathbb{R}^{n_s \times n_{t}}$ and $\hat{\textbf{S}}_s \in \mathbb{R}^{n_s \times n_{t}}$ are surrogate model predictions for pressure and saturation in the storage aquifer, $n_t = 10$ is the number of surrogate model time steps, and $\widetilde{\textbf{w}}$ represents the deep neural network parameters for the flow surrogate models determined during training. 

The original recurrent R-U-Net model developed by \citet{tang2022deep} treats 3D geomodels drawn from a single geological scenario. This surrogate model predicts saturation and pressure in the storage aquifer and surface vertical displacement above the storage aquifer. It involves a residual U-Net and a recurrent neural network, with the same network architecture used for pressure, saturation and surface displacement variables. All training runs with this approach are required to be fully coupled simulation runs. A direct extension of the method to treat multiple geological scenarios will thus be very expensive, since a large number of fully coupled simulation runs, spanning the range of possible metaparameters and realizations, will be required for training.

\begin{figure}[!ht]
\centering   
{\includegraphics[width = 152mm]{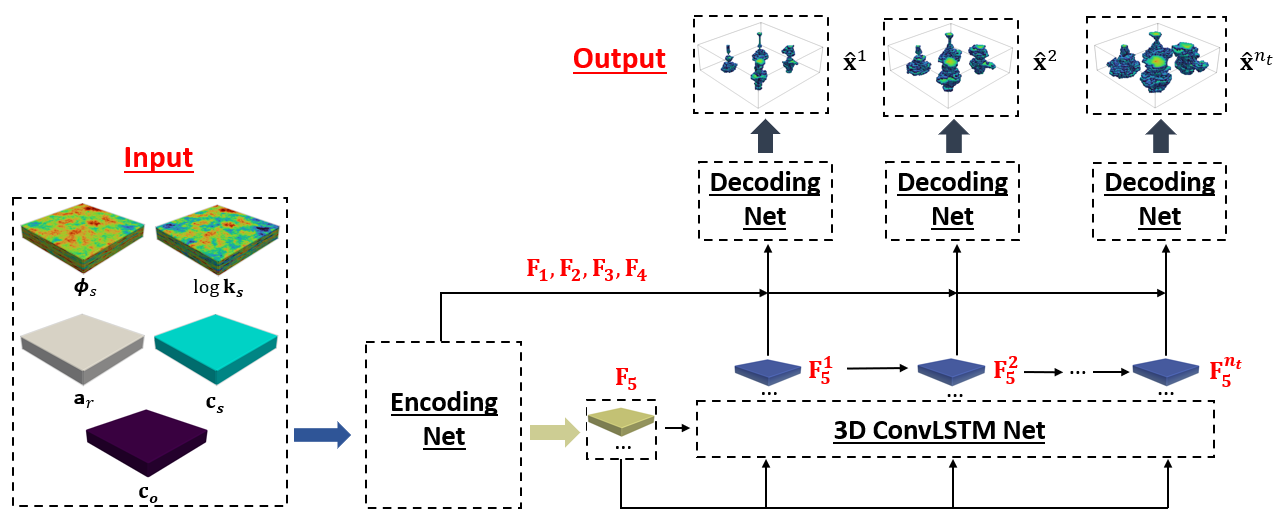}}
\caption{Schematic of the extended recurrent R-U-Net architecture for pressure and saturation prediction in the storage aquifer over $n_t$ time steps. Figure modified from \citet{han2023surrogate}.}
\label{nn_S_p}
\end{figure}

The recurrent R-U-Net model introduced by \citet{tang2022deep} was extended to treat geomodels drawn from multiple geological scenarios, for flow-only problems, by \citet{han2023surrogate}. The extended model involves three input channels, which correspond to the log-permeability field, the porosity field, and the permeability anisotropy ratio. In this study, two additional input channels are introduced into the recurrent R-U-Net architecture, namely the effective rock compressibility for the storage aquifer and the effective rock compressibility for the overburden. These quantities vary between realizations, though they are spatially constant within a particular realization. The input fields are nonetheless taken to be of the same dimensions as the other input channels. The general network architecture is shown in Fig.~\ref{nn_S_p}. Note it is not necessary to include effective rock compressibility for the underburden as an input channel because this quantity is closely correlated with effective rock compressibility for the overburden (see Table~\ref{rock physics}).

As noted earlier, there can be situations where the pressure or saturation fields from flow-only simulations with effective rock compressibility display sizeable errors relative to coupled flow-geomechanics results. In an earlier implementation, we used a single effective rock compressibility (rather than different $c_o$, $c_s$ and $c_u$) in the flow-only runs, and in that case we observed a mean pressure error of about 8.7\%. Through the use of a limited number of coupled simulation runs and the same network (Fig.~\ref{nn_S_p}), with an initial learning rate of 0.0001 rather than 0.0003 (used for flow-only runs), we were able to `fine-tune' the flow-only pressure solution. As a result, the mean error was reduced to about 1.1\%. An approach along these lines should be applicable for more challenging systems involving, e.g., fractures or faults.

\subsection{Surrogate model for surface displacement}
\label{sec:surr_displ}

A separate (new) network architecture is introduced to predict surface displacement. Consistent with \citet{tang2022deep}, the goal here is to predict vertical displacement at the Earth's surface in the region directly above the storage aquifer. This quantity is of interest because it can be measured by satellite (InSAR) and, as we will see, it is informative for history matching. We denote the surface vertical displacement field from the coupled flow-geomechanics simulations as $\textbf{d}_g \in \mathbb{R}^{n_g \times n_{ts}}$, where $n_g$ is the number of finite element nodes in the region of interest. To construct cell-based (rather than node-based) quantities, we average the four nodal displacements on the top surface of each cell~\citep{tang2022deep}. This quantity is denoted as $\textbf{d}_{gc} \in \mathbb{R}^{n_{gc} \times n_{ts}}$, where $n_{gc} = 80 \times 80=6400$ represents the number of cells in the top layer of the model above the storage aquifer.

A key observation is that surface displacement is strongly impacted by the pressure and saturation fields in the storage aquifer, along with the flow and geomechanical properties of the storage aquifer and overburden. Our strategy exploits these important relationships by using these fields as inputs to the surface displacement network. We can express this treatment as

\begin{equation} \label{coupled surrogate}
\hat{\textbf{d}}_{gc}^{t} = \hat{f}\left(\mathbf{m}_f, \hat{\textbf{p}}_{s}^t, \hat{\textbf{S}}_{s}^t, t; \textbf{w}\right),
\end{equation}
\noindent where $\hat{\textbf{d}}_{gc}^{t} \in \mathbb{R}^{n_{gc}}$ is the surrogate model prediction for the surface vertical displacement field directly above the storage aquifer at time $t$ (the same time steps as given above are considered), $\hat{\textbf{p}}_{s}^t$ and $\hat{\textbf{S}}_{s}^t$ are the extended recurrent R-U-Net surrogate model predictions for pressure and saturation in the storage aquifer at time $t$, and $\textbf{w}$ represents the deep neural network parameters for the surface displacement surrogate model determined during training. Because the pressure and saturation fields appearing in Eq.~\ref{coupled surrogate} are provided by the surrogate model, it is important that these be of high accuracy.

\begin{figure}[!ht]
\centering   
{\includegraphics[width = 152mm]{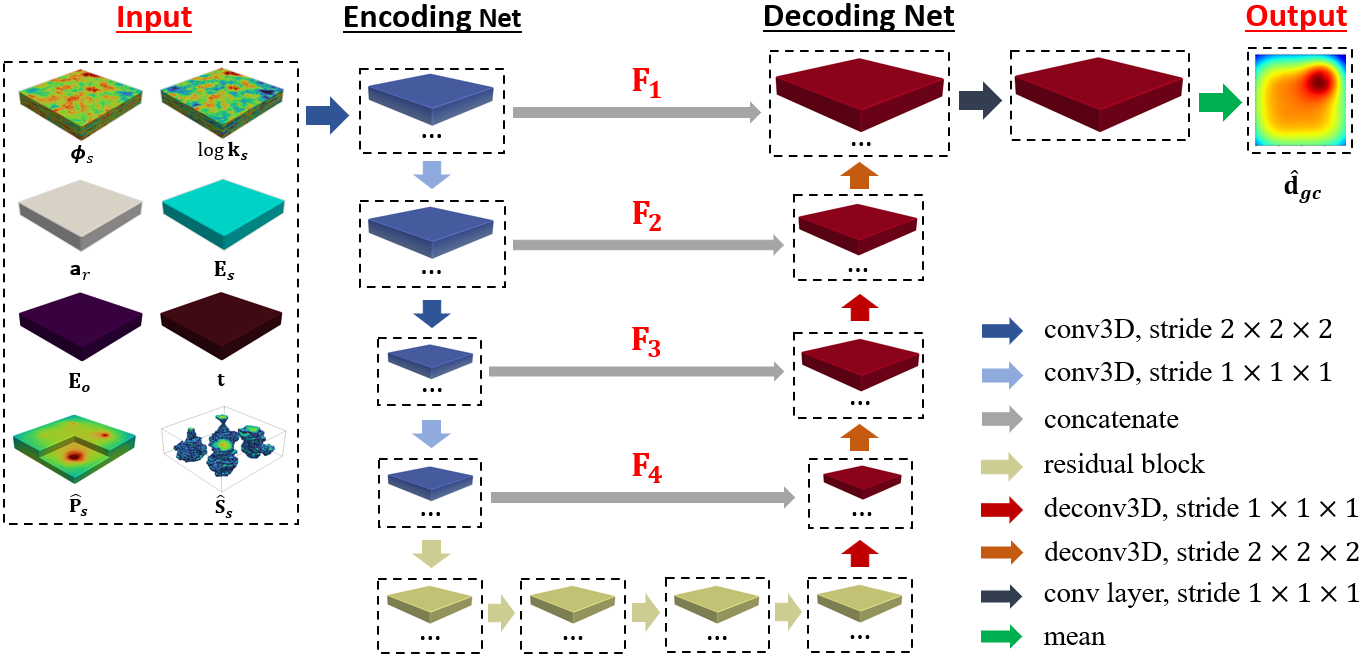}}
\caption{Schematic of the R-U-Net architecture for surface displacement prediction at a specific time.}
\label{nn_d}
\end{figure}

The residual U-Net (R-U-Net) model, shown in Fig.~\ref{nn_d}, is designed to predict surface displacement at a particular time. With this treatment, rather than use a recurrent neural network in the model architecture, dynamic quantities (aquifer pressure and saturation) and time are included as input channels and incorporated into the encoding network. This network design leads to improved prediction accuracy and a reduction in the number of runs required for training. Five additional input channels, which are similar to those in the extended recurrent R-U-Net for pressure and saturation prediction, are used to characterize the geomodel. In the surface displacement network, however, we input Young's modulus for the storage aquifer ($E_s$) and overburden ($E_o$) instead of effective rock compressibilities (as in Fig.~\ref{nn_S_p}).

Consistent with the pressure and saturation surrogate model, all input channels, including time and Young's modulus, are taken to be of the same dimension. The encoding network downsamples all eight input channels simultaneously to generate a sequence of latent feature maps at different levels of fidelity, which is intended to capture the multiscale characteristics of the problem. These multifidelity latent feature maps are incorporated into the decoding network, where they are combined with upsampled feature maps to provide predictions for surface displacement. Details of the model architecture are provided in Table~\ref{nn-architecture}. The last convolutional layer of the R-U-Net model generates a 3D  prediction, which is then averaged vertically to provide a 2D surface displacement map. This treatment is consistent with the approach used by \citet{tang2022deep}. 

We considered other approaches for directly predicting the 2D fields. These included the addition of a max pooling (or mean pooling) layer after the last convolutional layer to transform the 3D field into a 2D representation, followed by 2D convolutional and residual layers. We did not achieve improved accuracy with these network modifications. Network architectures that differ significantly from that Fig.~\ref{nn_d} and Table~\ref{nn-architecture} could also be considered for surface displacement prediction. It is possible, for example, that the time rollout strategy used by \citet{tang2024graph}, where time is not treated as an input channel, might lead to improved accuracy or reduced training demands. It will be useful to compare different network architectures in future work. 

\begin{table}[!ht]
\footnotesize
\begin{center}
\caption{Architecture of the R-U-Net model for surface displacement prediction}
\label{nn-architecture}
\renewcommand{\arraystretch}{1.15} 
\begin{tabular}{ c c c } 
\hline
\textbf{Network} & \textbf{Layer}  & \textbf{Output}\\ 
\hline
Encoder & Input & (n$_{x}$, n$_{y}$, n$_{z}$, 8) \\ 
 & conv, 16 filters of size $3 \times 3 \times 3$, stride 2 & ($\frac{n_x}{2}, \frac{n_y}{2}, \frac{n_z}{2}$, 16) \\ 
 & conv, 32 filters of size $3 \times 3 \times 3$, stride 1 & ($\frac{n_x}{2}, \frac{n_y}{2}, \frac{n_z}{2}$, 32) \\
 & conv, 32 filters of size $3 \times 3 \times 3$, stride 2 & ($\frac{n_x}{4}, \frac{n_y}{4}, \frac{n_z}{4}$, 32) \\
 & conv, 64 filters of size $3 \times 3 \times 3$, stride 1 & ($\frac{n_x}{4}, \frac{n_y}{4}, \frac{n_z}{4}$, 64) \\
 & residual block, 64 filters of size $3 \times 3 \times 3$, stride 1 & ($\frac{n_x}{4}, \frac{n_y}{4}, \frac{n_z}{4}$, 64) \\
 & residual block, 64 filters of size $3 \times 3 \times 3$, stride 1 & ($\frac{n_x}{4}, \frac{n_y}{4}, \frac{n_z}{4}$, 64) \\
\hline
Decoder & residual block, 64 filters of size $3 \times 3 \times 3$, stride 1 & ($\frac{n_x}{4}, \frac{n_y}{4}, \frac{n_z}{4}$, 64) \\
 & residual block, 64 filters of size $3 \times 3 \times 3$, stride 1 & ($\frac{n_x}{4}, \frac{n_y}{4}, \frac{n_z}{4}$, 64) \\
 & deconv, 64 filters of size $3 \times 3 \times 3$, stride 1 & ($\frac{n_x}{4}, \frac{n_y}{4}, \frac{n_z}{4}$, 64) \\
 & deconv, 32 filters of size $3 \times 3 \times 3$, stride 2 & ($\frac{n_x}{2}, \frac{n_y}{2}, \frac{n_z}{2}$, 32) \\
 & deconv, 32 filters of size $3 \times 3 \times 3$, stride 1 & ($\frac{n_x}{2}, \frac{n_y}{2}, \frac{n_z}{2}$, 32) \\
 & deconv, 16 filters of size $3 \times 3 \times 3$, stride 2 & (n$_{x}$, n$_{y}$, n$_{z}$, 16) \\
 & conv layer, 1 filters of size $3 \times 3 \times 3$, stride 1 & (n$_x$, n$_y$, n$_z$, 1) \\
\hline
Output & output layer, mean computation over $z$ direction & (n$_x$, n$_y$, 1) \\
\hline
\end{tabular}
\end{center}
\end{table}

\subsection{Training procedures}
\label{sec:training}

Normalization is applied prior to the training process. 
Saturation in each cell is already between 0 and 1, so no normalization is required. A time-dependent normalization is applied for both pressure and surface displacement. The normalized pressure is computed using

\begin{equation} \label{normalization}
(\bar{p}_s)_{i,j}^t = \frac{(p_s)_{i,j}^t - \mathrm{mean}{((p_s)_{i,j})}}{\mathrm{std}{((p_s)_{i,j})}}, 
\end{equation}
\noindent where $(p_s)_{i,j}^t$ and $(\bar{p}_s)_{i,j}^t$ are the pressure and normalized pressure in cell $j$ in geomodel $i$, at time step $t$, and $\mathrm{mean}{((p_s)_{i,j})}$ and $\mathrm{std}{((p_s)_{i,j})}$ represent the mean and standard deviation of pressure, computed over all storage aquifer pressure fields at all time steps and all training runs. Similarly, normalized surface displacement is given by

\begin{equation} \label{displacement normalization}
(\bar{d}_g)_{i,j}^t = \frac{(d_g)_{i,j}^t - \mathrm{mean}{((d_g)_{i,j})}}{\mathrm{std}{((d_g)_{i,j}})}, 
\end{equation}
\noindent where $(d_g)_{i,j}^t$ and $(\bar{d}_g)_{i,j}^t$ are the surface displacement and normalized surface displacement, and $\mathrm{mean}{((d_g)_{i,j})}$ and $\mathrm{std}{((d_g)_{i,j})}$ are the mean and standard deviation of surface displacement, computed over all surface displacement fields at all time steps and all training runs. These normalizations were found to result in slightly better prediction accuracy than other commonly used approaches. 

The training procedure for the pressure and saturation surrogate models entails the minimization of the $L_2$ error between simulated and predicted pressure and saturation, as described by \citet{han2023surrogate}. Different numbers of training samples will be considered in Section~\ref{Surrogate-performance}. For the largest number of samples (4000), the training procedure for each network requires about 16~hours using a Nvidia A100 GPU. This corresponds to 300~epochs, a batch size of 4, and an initial learning rate of 0.0003. Training times are reduced when fewer samples are considered.

The training objective for the surface displacement surrogate model can be expressed as

\begin{equation} \label{minimization}
\textbf{w}^{\ast} = \operatorname*{argmin}_{\textbf{w}} \frac{1}{n_{smp}} \sum_{i=1}^{n_{smp}}\|\hat{\textbf{x}}_{i} - \textbf{x}_{i}\|_2^2,
\end{equation}
\noindent where $\textbf{w}^{\ast}$ denotes the optimal network parameters, $n_{smp}$ is the number of training samples used in the surface displacement surrogate model, and $\textbf{x}_{i}$ and $\hat{\textbf{x}}_{i}$ represent the normalized surface displacement from the coupled flow-geomechanics simulation and the surrogate model, for training sample $i$, respectively. The performance of this network will be evaluated for $n_{smp}=$ 2000 and 4000. These cases correspond to 200 and 400 coupled flow-geomechanics simulation runs, with $n_t = 10$ time steps. Training for the surface displacement network (with 4000 training samples) requires only about 3~hours on the same Nvidia A100 GPU. This training corresponds to 300~epochs, a batch size of 16, and an initial learning rate of 0.0003.

\section{Surrogate Model Evaluation} 
\label{Surrogate-performance}
In this section, we first evaluate surrogate model performance for varying numbers of training samples. Results using two different approaches to predict surface displacement will be compared. Detailed CO$_2$ saturation and surface displacement fields for highly heterogeneous geomodels will then be presented. Finally, comparisons in terms of ensemble statistics for surface displacement will be provided.

\subsection{Surrogate model error statistics}
\label{sec:error_statistics}

We first generate a total of 4000 geomodel realizations, of the full domain, to be used for training. Flow-only simulation, with the effective rock compressibility for each model domain computed using Eq.~\ref{compressibility}, is performed for each realization. The pressure and saturation fields in the storage aquifer at $n_t= 10$ time steps are collected as training data for the flow surrogate model. Next, 400 new geomodels of the full domain are generated, and a coupled flow-geomechanics simulation is performed for each realization. The surface displacement fields directly above the storage aquifer at $n_t = 10$ time steps are collected to train the surface displacement surrogate model.

In terms of runtimes, a high-fidelity GEOS flow-only run, using 32 AMD EPYC-7543 CPU cores in parallel, takes about 8~minutes. A coupled flow-geomechanics GEOS run, also using 32 AMD EPYC-7543 CPU cores, requires about 120~minutes. A surrogate model evaluation for saturation, pressure, and surface displacement (at 10 time steps) requires only about 0.15~second per run on a single Nvidia A100 GPU. This runtime speedup is very dramatic, but the training simulations and network training require substantial computation.

The set of $n_e = 500$ new geomodels applied in Section~\ref{Comparison} to assess the accuracy of the effective rock compressibility treatment constitute the test set, which will now be used to evaluate surrogate model performance. As noted earlier, both coupled flow-geomechanics  and flow-only (with effective rock compressibility) simulations are performed on these geomodels. From these results, we compute surrogate model relative errors for CO$_2$ saturation, pressure, and surface displacement. For saturation and pressure, these error quantities, for test sample $i$ ($i = 1, \dots, 500$), are given by

\begin{equation} \label{surr_error_s}
\delta_S^i = \frac{1}{n_sn_t} \sum_{j=1}^{n_s} \sum_{t=1}^{n_t} \frac{| (\hat{S}_s)_{i,j}^t - (S_s)_{i,j}^t |}{(S_s)_{i,j}^t + \epsilon}, 
\end{equation}
\begin{equation} \label{surr_error_p}
\delta_p^i = \frac{1}{n_sn_t} \sum_{j=1}^{n_s} \sum_{t=1}^{n_t} \frac{| (\hat{p}_s)_{i,j}^t - (p_s)_{i,j}^t |}{(p_s)_{i,\mathrm{max}}^t - (p_s)_{i,\mathrm{min}}^t}, 
\end{equation}
where $(\hat{S}_s)_{i,j}^t$ and $(\hat{p}_s)_{i,j}^t$ are CO$_2$ saturation and pressure predictions from the flow surrogate model, for test case $i$, storage aquifer cell $j$, and surrogate model output time step $t$. The relative error for surface displacement for sample $i$ is defined as

\begin{equation} \label{surr_error_d}
\delta_d^i = \frac{1}{n_{gc}n_t} \sum_{j=1}^{n_{gc}} \sum_{t=1}^{n_t} \frac{| (\hat{d}_g)_{i,j}^t - (d_g)_{i,j}^t |}{(d_g)_{i,j}^t},
\end{equation}
where $n_{gc}$ = 6400 is the number of cells on the surface above the storage aquifer, and $(d_g)_{i,j}^t$ and $(\hat{d}_g)_{i,j}^t$ are surface displacement predictions from the coupled flow-geomechanics simulation and surrogate model, respectively, for test case $i$, cell $j$, and time step $t$. 

\begin{figure}[!ht]
\centering   
\subfloat[Saturation relative error]{\label{error_s}\includegraphics[width = 85mm]{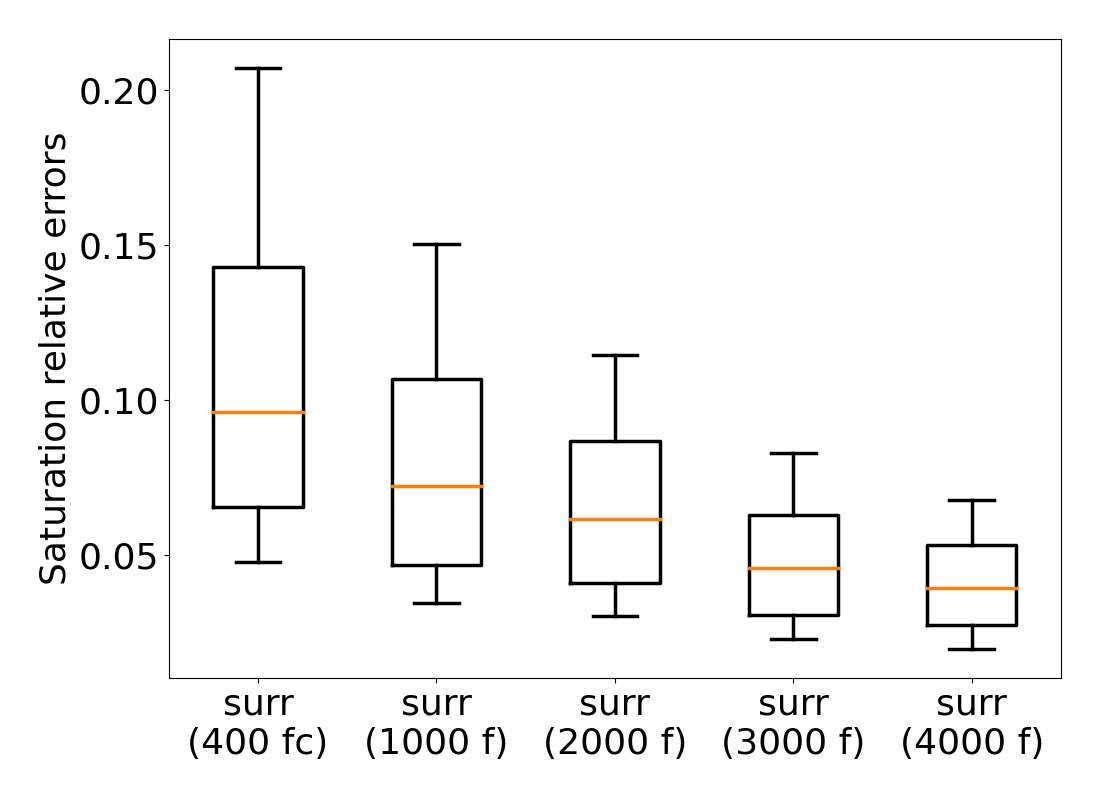}}
\hspace{1mm}
\subfloat[Pressure relative error]{\label{error_p}\includegraphics[width = 85mm]{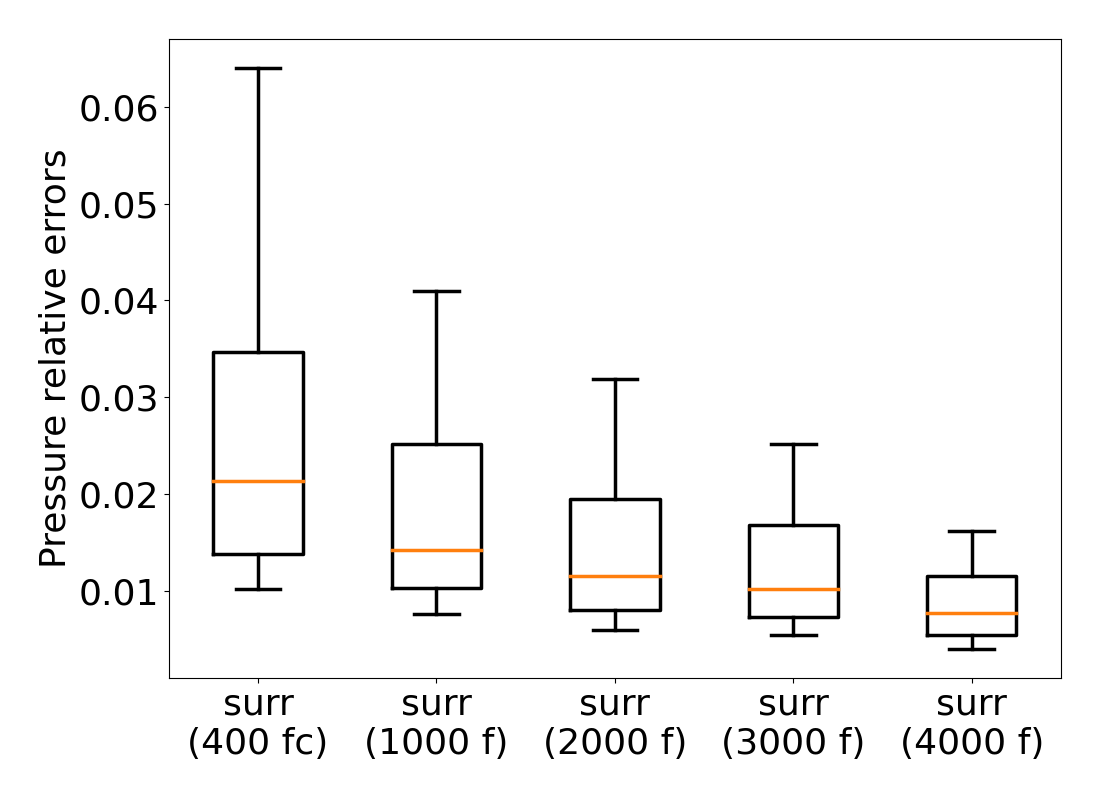}} \\
\caption{Saturation and pressure relative errors for test set of 500 coupled flow-geomechanics cases. Results shown for surrogate model trained with 400 fully coupled flow-geomechanics runs (400~fc), 1000 flow-only runs (1000~f), 2000 flow-only runs (2000~f), 3000 flow-only runs (3000~f), and 4000 flow-only runs (4000~f). Boxes display P$_{90}$, P$_{75}$, P$_{50}$, P$_{25}$ and P$_{10}$ errors.}
\label{errors_box}
\end{figure}

\begin{figure}[!ht]
\centering  
\includegraphics[width=16.2cm]{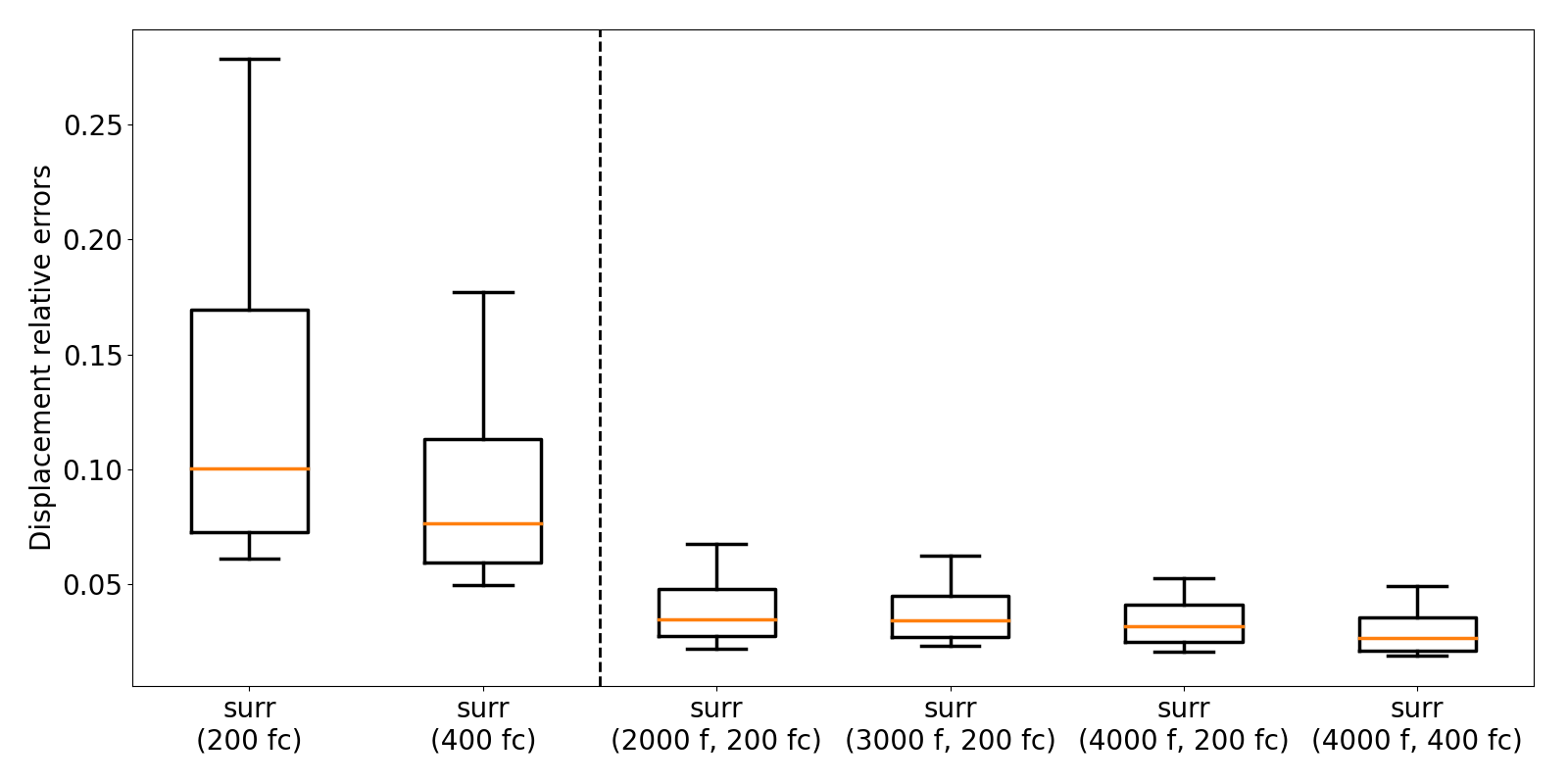}
\caption{Surface displacement relative errors for test set of 500 coupled flow-geomechanics cases. Results to the left of the vertical line are for a surrogate model based on the network in Fig.~\ref{nn_S_p} and trained with 200 and 400 fully coupled (fc) runs. Results to the right of the vertical line are for the network in Fig.~\ref{nn_d} and involve flow-only (f) and fully coupled runs. Boxes display P$_{90}$, P$_{75}$, P$_{50}$, P$_{25}$ and P$_{10}$ errors.}
\label{errors_box_d}
\end{figure}

The test-case relative errors for saturation, pressure, and surface displacement are presented as box plots in Figs.~\ref{errors_box} and~\ref{errors_box_d}. In these box plots, the P$_{25}$ and P$_{75}$ (25th and 75th percentile) relative errors are indicated by the bottom and top of each box, while the solid red line represents the P$_{50}$ (median) relative error. The lines extending beyond the boxes show the P$_{10}$ and P$_{90}$ relative errors.

The relative errors for saturation and pressure trained with different numbers of simulation runs are displayed in Fig.~\ref{errors_box}. The mean and median relative errors for the saturation surrogate model, trained with simulation results from 400 fully coupled flow-geomechanics runs (400~fc), are 11.3\% and 9.6\%, respectively. For the pressure surrogate model, these relative errors are 3\% and 2.1\%. The relatively large saturation errors suggest that more training samples are required. The extremely small differences between the coupled and flow-only (with effective compressibility) simulations allows us to use flow-only simulations for this training. This provides significant computational savings since the flow-only runs are $15\times$ faster than the coupled runs. In Fig.~\ref{errors_box}a, we see that the median relative error for saturation decreases from 7.2\% for 1000 samples, to 6.1\% for 2000 samples, to 4.6\% for 3000 samples, and to 3.9\% for 4000 samples. For pressure, the median relative error drops to 0.8\% with 4000 samples. The saturation and pressure surrogate models trained using 4000 flow-only runs are used in all subsequent predictions in Section~\ref{sec:surrogate_predictions}, and for history matching in Section~\ref{History-matching}.

For surface displacement prediction, we compare two network architectures and different numbers of training samples. Results with the approach in \citet{tang2022deep}, which entails using the same network architecture for surface displacement as is used for saturation and pressure (Fig.~\ref{nn_S_p}), are shown to the left of the vertical dashed line in Fig.~\ref{errors_box_d}. The training runs in these cases entail coupled flow-geomechanics simulations. The mean and median relative errors are 14.1\% and 10\% using 200 training samples, and 9.9\% and 7.7\% using 400 samples. These errors can be reduced by using a larger  number of training samples, but this is computationally expensive if coupled runs are performed (the errors in \citet{tang2022deep}, where 2000 coupled training runs were used and the overall setup was considerably simpler, are much smaller).

Results using the approach proposed in this paper, in which the large majority of training runs are flow-only simulations and the network in Fig.~\ref{nn_d} is used, are shown to the right of the vertical dashed line in Fig.~\ref{errors_box_d}. The mean and median relative errors are 4.2\% and 3.5\% using 2000 flow-only (for the pressure and saturation surrogate models) and 200 coupled runs, and 4\% and 3.4\% using 3000 flow-only and 200 coupled runs. With 4000 flow-only and 200 coupled runs, the mean and median relative errors are 3.6\% and 3.2\%, and with 4000 flow-only and 400 coupled runs, the errors decrease to 3.2\% and 2.6\%. These results are significantly more accurate than those using only 400 coupled runs, demonstrating the advantages, in terms of efficiency and accuracy, of our proposed framework. All subsequent surrogate model and history matching results involving surface displacements apply this framework, trained with 4000 flow-only and 400 coupled flow-geomechanics runs. 

Recall that the coupled simulation runs require $15\times$ the computational time of the flow-only runs. Thus, the overall time for training runs for the 400~fc case in Fig.~\ref{errors_box_d} is about the same as that for the 3000~f,~200~fc case (3000/15+200=400). The substantial advantage in terms of prediction accuracy using the latter approach clearly demonstrates the benefit of our treatments. The 2000~f,~200~fc case requires the equivalent of about 333 coupled runs, which is less than that for the 400~fc case. The enhanced accuracy provided by the 2000~f,~200~fc case illustrates that our approach is not overly sensitive to the exact number of training runs of either type.

\subsection{Surrogate predictions for individual geomodels}
\label{sec:surrogate_predictions}
 
\begin{figure}[H]
\centering   
\subfloat[Realization~1 ($\mu_{\log k}$=3.51, $\sigma_{\log k}$=1.94, $a_r$=0.04)]{\label{logk_1}\includegraphics[width = 49mm]{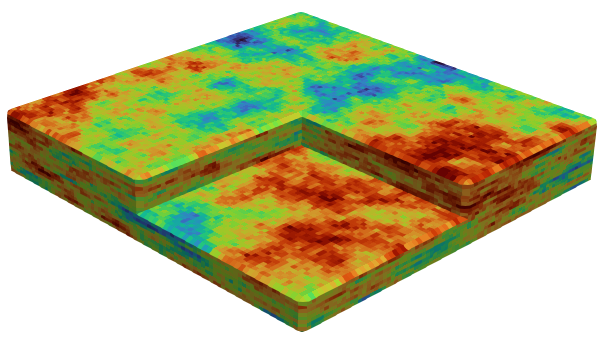}}
\hspace{3mm}
\subfloat[Realization~2 ($\mu_{\log k}$=3.12, $\sigma_{\log k}$=2.09, $a_r$=0.02)]{\label{logk_2}\includegraphics[width = 49mm]{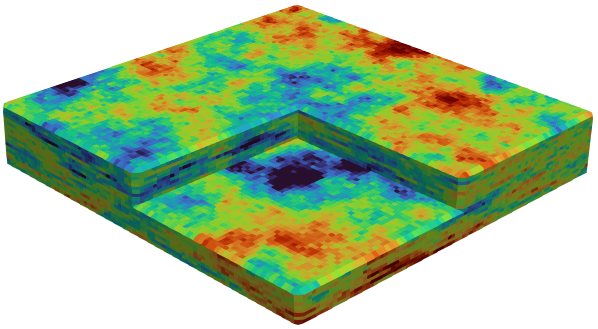}}
\hspace{3mm}
\subfloat[Realization 3 ($\mu_{\log k}$=2.83, $\sigma_{\log k}$=2.44, $a_r$=0.03)]{\label{logk_3}\includegraphics[width = 49mm]{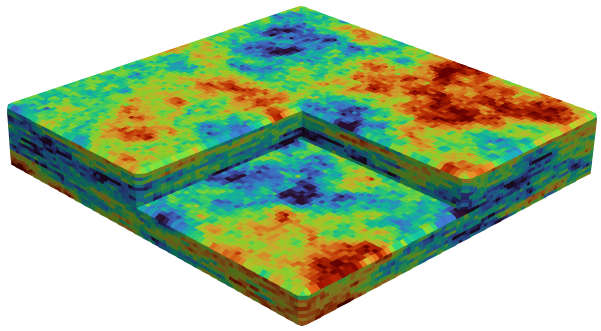}}
\includegraphics[width = 7.5mm]{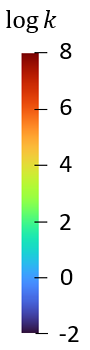}
\caption{Log-permeability fields ($\log k_x$, with $k_x$ in md, is shown)  for three realizations. Geomodel in (a) is the `true' model used for history matching in Section~\ref{`true' model 1}.}
\label{True_Models}
\end{figure}

We now present detailed saturation and surface displacement fields for three particular geomodels. These realizations are shown in Fig.~\ref{True_Models}. Each has a high degree of permeability anisotropy ($k_z/k_x \leq 0.04$) and substantial permeability variability ($\sigma_{\log k} \geq 1.94$), which suggest the saturation fields will be complex and irregular. The relative saturation errors for all three geomodels are larger than the median relative error, indicating these are challenging cases.

\begin{figure}[H]
\centering   
\subfloat[Realization 1 (sim)]{\label{sim_1}\includegraphics[width=45mm]{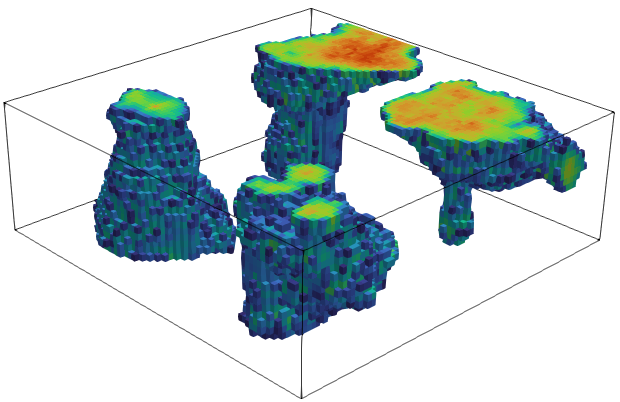}}
\hspace{3mm}
\subfloat[Realization 2 (sim)]{\label{sim_2}\includegraphics[width=45mm]{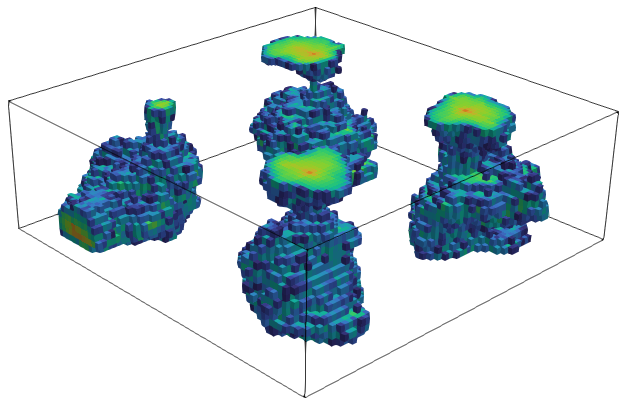}}
\hspace{3mm}
\subfloat[Realization 3 (sim)]{\label{sim_3}\includegraphics[width=45mm]{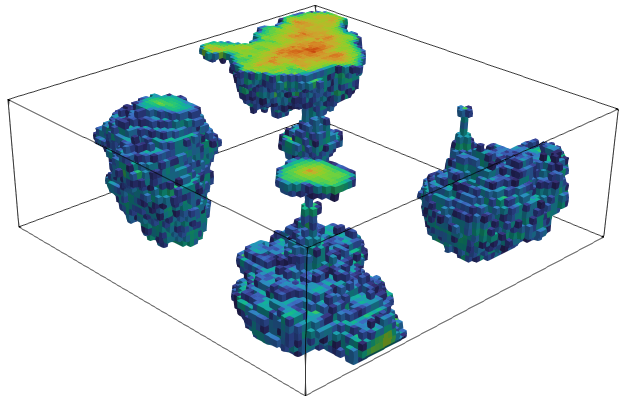}}
\includegraphics[width=9mm]{Figure/Surrogate/Sw_Scale.png}\\[1ex]
\subfloat[Realization 1 (surr)]{\label{surr_1}\includegraphics[width=45mm]{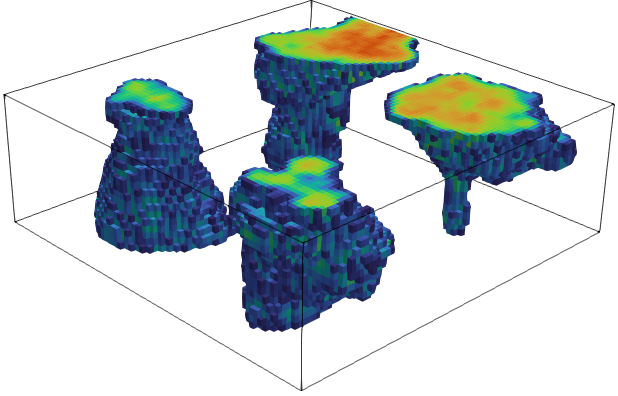}}
\hspace{3mm}
\subfloat[Realization 2 (surr)]{\label{surr_2}\includegraphics[width=45mm]{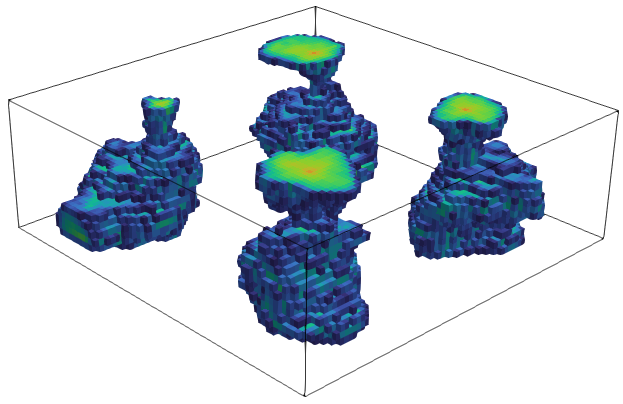}}
\hspace{3mm}
\subfloat[Realization 3 (surr)]{\label{surr_3}\includegraphics[width=45mm]{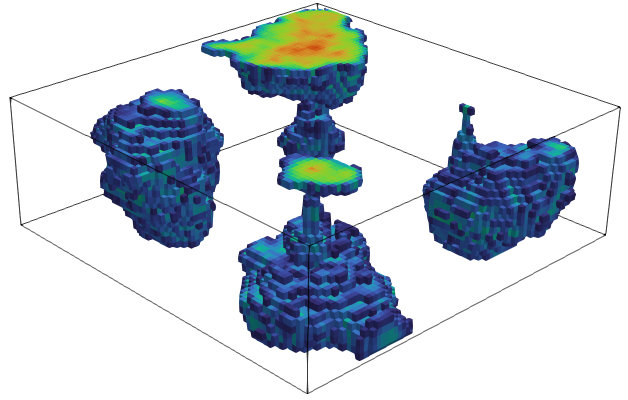}}
\includegraphics[width=9mm]{Figure/Surrogate/Sw_Scale.png}\\[1ex]
\caption{CO$_2$ saturation in the storage aquifer at the end of injection (30~years) from GEOS coupled flow-geomechanics simulation (upper row) and the surrogate model (lower row) for the three highly heterogeneous geomodels shown in Fig.~\ref{True_Models}.}
\label{S:saturation}
\end{figure}

The CO$_2$ saturation plumes at the end of the injection period (30~years) for the three models are shown in Fig.~\ref{S:saturation}. The upper row displays the reference coupled flow-geomechanics simulation results, and the lower row shows the surrogate model predictions. The two sets of results are in close visual agreement in all cases. In all three realization, one or more plumes exhibit highly irregular shapes, and these are well represented by the surrogate model. Note that some plumes, especially in realizations~1 and 3, migrate out of the storage aquifer, and this behavior is also captured by the surrogate model. Although detailed comparisons are not shown in this paper, we also observe close agreement between simulation and surrogate model results for pressure, consistent with the very low relative errors in Fig.~\ref{errors_box}b.

\begin{figure}[H]
\centering   
\subfloat[Realization 1 (sim)]{\label{sim_d_1_2_years}\includegraphics[width=50mm]{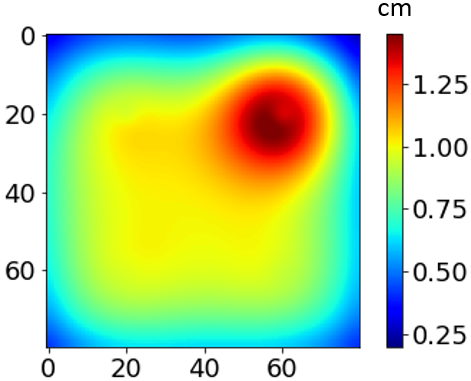}}
\hspace{7mm}
\subfloat[Realization 2 (sim)]{\label{sim_d_2_2_years}\includegraphics[width=48mm]{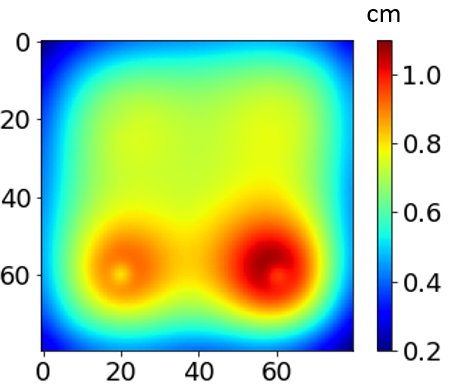}}
\hspace{7mm}
\subfloat[Realization 3 (sim)]{\label{sim_d_3_2_years}\includegraphics[width=48mm]{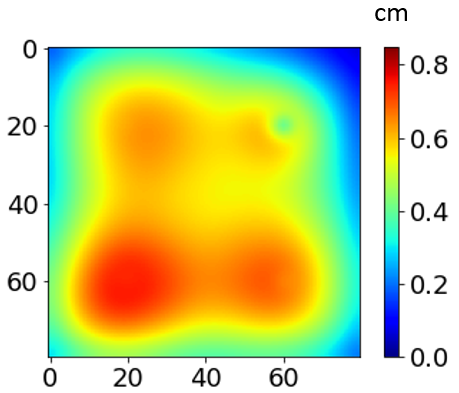}}\\[1ex]
\subfloat[Realization 1 (surr)]{\label{surr_d_1_2_years}\includegraphics[width=50mm]{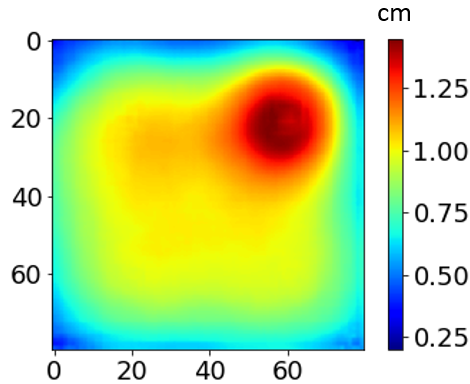}}
\hspace{7mm}
\subfloat[Realization 2 (surr)]{\label{surr_d_2_2_years}\includegraphics[width=48mm]{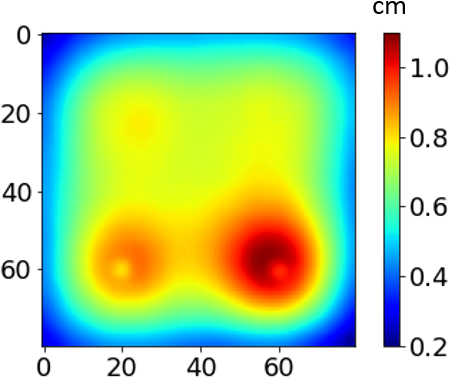}}
\hspace{7mm}
\subfloat[Realization 3 (surr)]{\label{surr_d_3_2_years}\includegraphics[width=48mm]{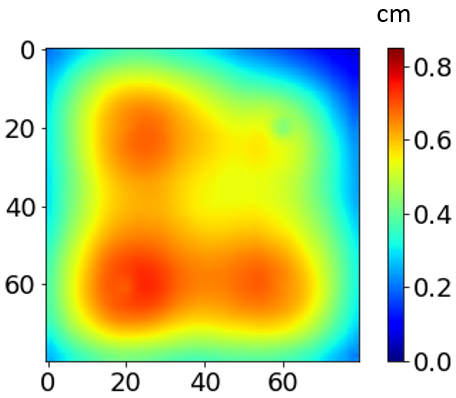}}\\[1ex]
\caption{Surface vertical displacement directly above the storage aquifer at 2~years from GEOS coupled flow-geomechanics simulation (upper row) and the surface displacement surrogate model (lower row) for the three geomodels shown in Fig.~\ref{True_Models}.}
\label{displacement_2_years}
\end{figure}

\begin{figure}[H]
\centering   
\subfloat[Realization 1 (sim)]{\label{sim_d_1}\includegraphics[width=48mm]{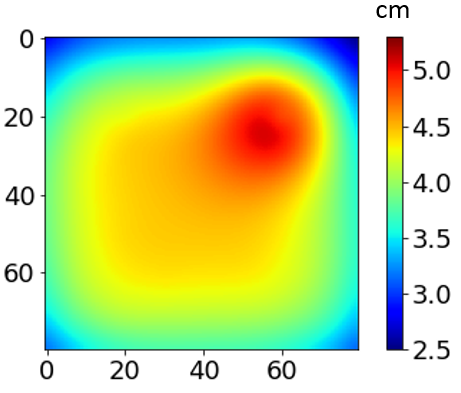}}
\hspace{7mm}
\subfloat[Realization 2 (sim)]{\label{sim_d_2}\includegraphics[width=48mm]{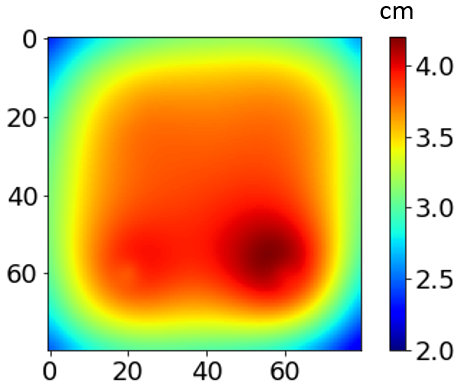}}
\hspace{7mm}
\subfloat[Realization 3 (sim)]{\label{sim_d_3}\includegraphics[width=48mm]{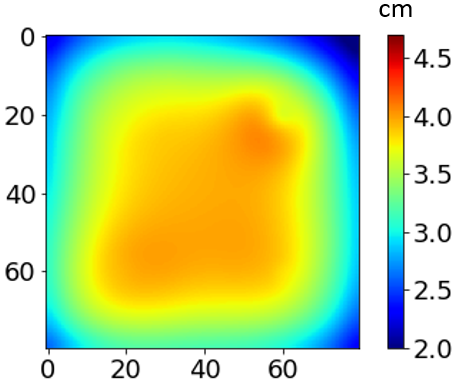}}\\[1ex]
\subfloat[Realization 1 (surr)]{\label{surr_d_1}\includegraphics[width=48mm]{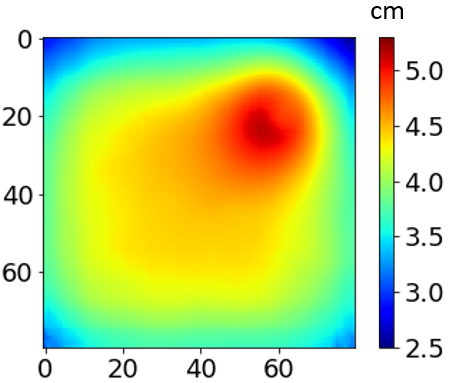}}
\hspace{7mm}
\subfloat[Realization 2 (surr)]{\label{surr_d_2}\includegraphics[width=48mm]{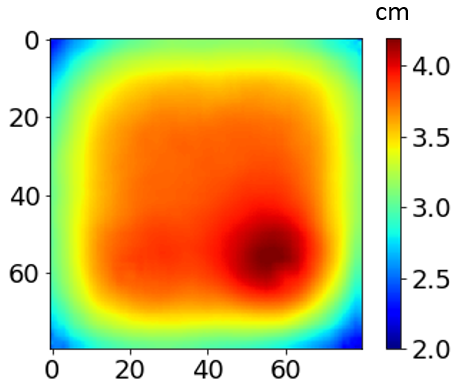}}
\hspace{7mm}
\subfloat[Realization 3 (surr)]{\label{surr_d_3}\includegraphics[width=48mm]{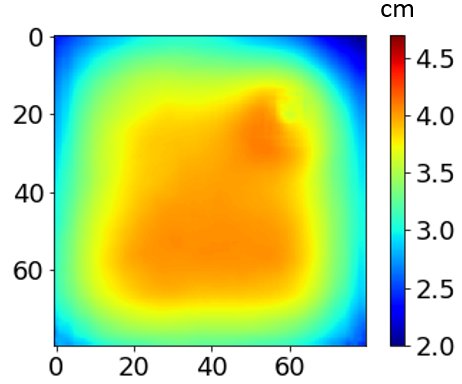}}\\[1ex]
\caption{Surface vertical displacement directly above the storage aquifer at 30~years from GEOS coupled flow-geomechanics simulation (upper row) and the surface displacement surrogate model (lower row) for the three geomodels shown in Fig.~\ref{True_Models}.}
\label{displacement_30_years}
\end{figure}

The surface displacement relative errors for realizations~1-3 are 1.7\%, 2.5\% and 5.1\%, respectively (median error over the full test set is 2.6\%). The surface displacement fields at early time (2~years) and at the end of the injection period (30~years) for the three geomodels are shown in Figs.~\ref{displacement_2_years} and~\ref{displacement_30_years}. The correspondence between the coupled flow-geomechanics simulation and surrogate model results is again very close, both at early and late time. The solution character differs considerably between the three cases. For example, realization~3 shows a smoother, more symmetric displacement field at 30~years, while realization~1 has noticeably more displacement in the upper right of the model. In addition, the displacement field for realization~3 is less uniform at 2~years than at 30~years, whereas realization~1 displays similar variation at both times. All of these behaviors are accurately captured by the surrogate model.

\begin{figure}[H]
\centering
\subfloat[Surface displacement above I1]{\label{d_ensemble_1}\includegraphics[width = 72mm]{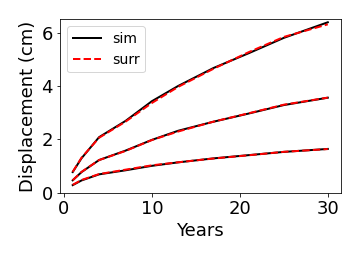}}
\hspace{2mm}
\subfloat[Surface displacement above I2]{\label{d_ensemble_2}\includegraphics[width = 72mm]{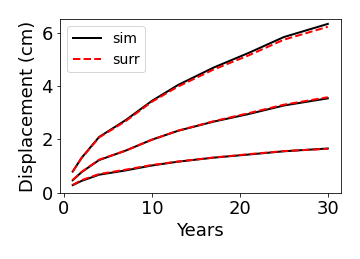}}
\\[1ex]
\subfloat[Surface displacement above I3]{\label{d_ensemble_3}\includegraphics[width = 72mm]{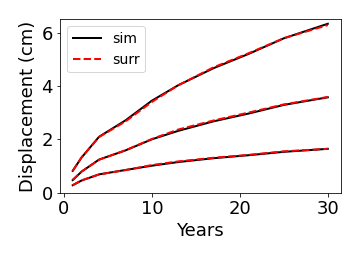}}
\hspace{2mm}
\subfloat[Surface displacement above I4]{\label{d_ensemble_4}\includegraphics[width = 72mm]{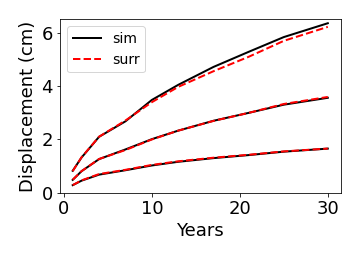}}
\\[1ex]
\caption{Surface displacement ensemble statistics from GEOS coupled simulation results (black solid curves) and surrogate model predictions (red dashed curves) at surface monitoring locations directly above the injection wells. The upper, middle and lower curves correspond to P$_{90}$, P$_{50}$ and P$_{10}$ results over the test set.}
\label{statistics_d}
\end{figure}

Finally, we present time series results for P$_{10}$, P$_{50}$ and P$_{90}$ surface displacement at four observation locations, for the simulation and surrogate models. These results are shown in Fig.~\ref{statistics_d}. The (surface) observation locations are directly above each of the four injection wells. In the figures, the solid black curves represent GEOS coupled flow-geomechanics simulation results, and the dashed red curves are surrogate model predictions. The upper, middle, and lower curves correspond to P$_{90}$, P$_{50}$ and P$_{10}$ results at the 10 time steps (results at different time steps may correspond to different models). Agreement is very close in all cases. Similar levels of agreement are also observed for surface displacement at other observation locations. Close correspondence in monitoring well statistics (specifically P$_{10}$, P$_{50}$, P$_{90}$ results for pressure and saturation) is also observed, though these results are not shown in this paper. 

\section{History Matching using Surrogate Model} 
\label{History-matching}
In this section, we apply the flow and surface displacement surrogate models within a hierarchical MCMC history matching procedure. We first describe the problem setup for history matching. A new approach to account for surrogate model error is then introduced. Next, we provide an overview of the MCMC-based data assimilation method. History matching results for a synthetic `true' model are then presented. Results that do not account for surrogate model error, along with results for a second true model (that is outside the prior range) are presented in Supplementary Information (SI). 

\subsection{Problem setup for history matching}
\label{History matching setup}

In a practical setting, the observations used for history matching correspond to data collected in the field. In this study, the data derive from coupled flow-geomechanics simulation results for a synthetic randomly sampled true model (denoted $\textbf{m}_{\mathrm{true}}$). We use simulation results, at both subsurface and surface observation locations, at five time steps (1, 2, 4, 6, and 9~years after the start of injection). The subsurface data include saturation at all 20~layers in the storage aquifer at the four observation wells O1--O4 (shown in Fig.~\ref{wells}). Pressure is assumed to be measured in all 20~layers of the storage aquifer in observation well O5. Thus we have a total of $5 \times 20 \times 4=400$ saturation measurements and 100 pressure measurements.

Note that the distance between any saturation observation well (O1–-O4) and the nearest injection well is 335~m. This is much less than the distance between the pressure observation well O5 and the injectors (4243~m). This setup results in a large amount of informative saturation data, especially at early time when the plumes have not progressed very far. Pressure effects are `felt' throughout the domain, so it is reasonable to use a single pressure observation well placed at the center of the storage aquifer. 

Vertical surface displacements, which can be measured by satellite (InSAR), are assumed to be available at 81 ground locations. These measurement locations span the entire region directly above the storage aquifer. They are arranged on a $9 \times 9$ grid, with the locations spaced every 1.5~km in the $x$ and $y$ directions, as shown in Fig.~\ref{observations}b. We thus have $5\times81=405$ surface uplift measurements, leading to a total of $n_m = 905$ surface and subsurface measurements.

Both measurement error and model error enter into the history matching procedure. Measurement error is associated with the precision of the measurement device and data interpretation. Model error in this work is associated with inaccuracy in the surrogate model relative to high-fidelity (GEOS) simulation. Following \citet{jiang2024history}, the standard deviation of measurement error is set to 0.02 saturation units and 0.095~MPa for pressure. For surface displacement, we use a standard deviation of 0.1~cm, which corresponds to the resolution of InSAR measurement for surface uplift~\citep{smittarello2022pair}. The observed data with measurement error, $\textbf{d}_{\mathrm{obs}}$, can be expressed as

\begin{equation} \label{noise}
\textbf{d}_{\mathrm{obs}} = \textbf{d}_{\mathrm{true}} + \bm{\epsilon} = f(\textbf{m$_{\mathrm{true}}$}) + \bm{\epsilon},
\end{equation}
\noindent where $f(\textbf{m$_{\mathrm{true}}$})$ = $\textbf{d}_{\mathrm{true}} \in \mathbb{R}^{n_m}$ represents the true data, which in our case are coupled flow-geomechanics simulation results at the observation locations for true model $\textbf{m}_{\mathrm{true}}$, and $\bm{\epsilon}$ denotes the measurement error. Here $\bm{\epsilon}$ is of $\textbf{0}$ mean and measurement error covariance $C_{\mathrm{D}}  \in \mathbb{R}^{n_m \times n_m}$. Measurement errors are assumed to be uncorrelated, so $C_{\mathrm{D}}$ is a diagonal matrix whose elements are the variance of measurement error~\citep{alfonzo2020seismic}.

\subsection{Treatment of model error}
\label{sec:model_error}

Model error can derive from many sources, including grid resolution effects, error from reduced or uncertain physics, etc. The covariance from these different sources can be combined, so it is possible to include multiple effects. Many approaches have been developed to treat model error in the context of history matching. \citet{oliver2018calibration} developed an iterative workflow for estimating total error covariance and calibrating models using an ensemble-based history matching method. The total error covariance, determined based on the residual of observation and history matched results, included both measurement error and model error covariance. \citet{rammay2019quantification} applied a joint inversion approach within an ensemble-based data assimilation framework to estimate both reservoir parameters and model error parameters. Model error in this case was due to grid coarsening. \citet{jiang2021treatment} also considered model error due to grid resolution. They incorporated a model error treatment into a data-space inversion procedure. In later work, \citet{jiang2024history} applied a grid refinement assessment to estimate grid resolution error. \citet{neto2021assimilating} developed an iterative workflow with an ensemble-based history matching method for cases with time-lapse seismic data. Both the calibrated model and the total error covariance were updated during iteration.

Despite the increasing use of surrogate models for data assimilation in subsurface flow applications, there do not appear to be many studies that have addressed the treatment of model error due to inaccuracy in surrogate  predictions. A few heuristic approaches have, however, been proposed. \citet{seabra2024ai}, for example, developed a hybrid data assimilation workflow based on ESMDA that used both surrogate modeling and high-fidelity simulation. They used high-fidelity simulation in the first ESMDA step and surrogate modeling for other steps. They also applied high-fidelity simulation to compute posterior results based on the final ESMDA parameters. This latter treatment was also applied by \citet{tang2022deep}. \citet{han2023surrogate} estimated overall surrogate model error statistics based on the average and standard deviation of saturation and pressure over the full ensemble of test cases, at all observation locations and time steps. With this approach, error was found to be unbiased (mean near zero), and the covariance matrix for surrogate model error contained only diagonal terms. Applying this procedure with the test set in this study, the mean and standard deviation of the overall surrogate model error are, respectively, 0.003 and 0.031 for saturation, 0.009~MPa and 0.067~MPa for pressure, and 0.007~cm and 0.105~cm for surface displacement. It is notable that we again observe error that is largely unbiased.

Here we apply a more formal approach for treating surrogate model error in history matching. Our procedure is as follows. Surrogate model error statistics can be computed using the high-fidelity coupled flow-geomechanics simulations for the test set. We compute a surrogate model error vector for each realization $i$ ($i = 1, \dots, 500$), denoted by $\bm{\epsilon}_{\mathrm{surr}}^i \in \mathbb{R}^{n_m}$, via

\begin{equation} \label{model_error}
\bm{\epsilon}_{\mathrm{surr}}^i = f(\textbf{m$_f^i$}) - \hat{f}(\textbf{m$_f^i$}) ,
\end{equation}
\noindent where $f(\textbf{m}_f^i)$ and $\hat{f}(\textbf{m}_f^i)$ represent the GEOS and surrogate model predictions for saturation, pressure, and surface displacement at the observation locations at all time steps for geomodel realization $\textbf{m}_f^i$. The surrogate model errors for the $n_e=500$ realizations are assembled into a centered matrix $D_{\mathrm{surr}} \in \mathbb{R}^{n_m \times n_e}$,

\begin{equation} \label{model_error_matrix}
D_{\mathrm{surr}} = \frac{1}{\sqrt{n_e - 1}} \left[\bm{\epsilon}_{\mathrm{surr}}^1 - \bar{\bm{\epsilon}}_{\mathrm{surr}} \quad \bm{\epsilon}_{\mathrm{surr}}^2 - \bar{\bm{\epsilon}}_{\mathrm{surr}} \quad \dots \quad \bm{\epsilon}_{\mathrm{surr}}^{n_e} - \bar{\bm{\epsilon}}_{\mathrm{surr}}\right],
\end{equation}
\noindent where $\bar{\bm{\epsilon}}_{\mathrm{surr}} \in \mathbb{R}^{n_m}$ is the mean of the surrogate model errors over the 500 test set realizations. The covariance of the surrogate model error is then given by

\begin{equation} \label{model_error_covariance_matrix}
C_{\mathrm{surr}} = D_{\mathrm{surr}}D_{\mathrm{surr}}^{T}.
\end{equation}

The covariance matrix $C_{\mathrm{surr}}$ computed using Eq.~\ref{model_error_covariance_matrix} is nondiagonal, and can thus represent errors that are spatially and temporally correlated. This is important in our problem because we have a large number of measurements, at different surface and subsurface locations and time steps, and the error correlation structure may be complicated. The covariance matrix $C_{\mathrm{surr}}$ is not truncated (e.g., by applying singular value decomposition). We note finally that a workflow that involves estimating surrogate model error covariance and performing history matching iteratively, such as that described by \citet{oliver2018calibration}, could be considered. Such an approach is not attempted here, however.

\subsection{Hierarchical MCMC sampling}
\label{MCMC sampling}

We now describe the history matching procedure used in this work. The basic approach entails the use of the noncentered preconditioned Crank-Nicolson within Gibbs algorithm introduced by \citet{chen2018dimension}. The specific implementation used here closely follows that in \citet{han2023surrogate}, which should be consulted for details.

Geomodels of the full domain are defined by the metaparameters $\boldsymbol{\uptheta}_{\mathrm{meta}}$ and PCA latent variables $\bm{\xi}$. The estimation of posterior distributions for both the metaparameters and associated realizations entails a hierarchical Bayes problem~\citep{chen2018dimension, xiao2021bayesian}. The probability density $\boldsymbol{\uptheta}_{\mathrm{meta}}$ and $\bm{\xi}$, conditioned to observation data $\bm{\mathrm{d}}_{\mathrm{obs}}$, is given by Bayes' theorem:

\begin{equation} \label{Bayes}
  p({\boldsymbol{\uptheta}_{\mathrm{meta}}, \bm{\xi} | \bm{\mathrm{d}}_{\mathrm{obs}}}) = \frac{p(\boldsymbol{\uptheta}_{\mathrm{meta}}, \bm{\xi}) p(\bm{\mathrm{d}}_{\mathrm{obs}}|{\boldsymbol{\uptheta}}_{\mathrm{meta}}, \bm{\xi})} {p(\bm{\mathrm{d}}_{\mathrm{obs}})}.
  \end{equation}
\noindent Here $p({\boldsymbol{\uptheta}_{\mathrm{meta}}, \bm{\xi}|\bm{\mathrm{d}}_{\mathrm{obs}}})$ is the posterior probability density function for the metaparameters and PCA latent variables, $p(\boldsymbol{\uptheta}_{\mathrm{meta}}, \bm{\xi})$ is the prior probability density function, ${p(\bm{\mathrm{d}}_{\mathrm{obs}})}$ enters the formulation as a normalization constant, and $p(\bm{\mathrm{d}}_{\mathrm{obs}} | {\boldsymbol{\uptheta}}_{\mathrm{meta}}, \bm{\xi})$ is the likelihood function. This function, which involves the mismatch between surrogate model predictions and observations, is given by

\begin{equation} \label{likelihood}
  p(\bm{\mathrm{d}}_{\mathrm{obs}} | {\boldsymbol{\uptheta}}_{\mathrm{meta}}, \bm{\xi}) = \\
  c \exp\left({{-\frac{1}{2}  \left(\bm{\mathrm{d}}_{\mathrm{obs}}-\hat{f}\left(\textbf{m}_f\left({\boldsymbol{\uptheta}}_{\mathrm{meta}}, \bm{\xi}\right)\right) \right)}^T C_{\mathrm{tot}}^{-1} \left(\bm{\mathrm{d}}_{\mathrm{obs}}-\hat{f}\left(\textbf{m}_f\left({\boldsymbol{\uptheta}}_{\mathrm{meta}}, \bm{\xi}\right)\right) \right)} \right),
\end{equation} 
\noindent where $c$ is a normalization constant, $\hat{f}\left(\textbf{m}_f\left({\boldsymbol{\uptheta}}_{\mathrm{meta}},  \bm{\xi}\right)\right)$ represents the surrogate model prediction for the observed quantities, and $C_{\mathrm{tot}}$ is the total error covariance given by

\begin{equation} \label{covariance}
  C_{\mathrm{tot}} = C_{\mathrm{D}} + C_{\mathrm{surr}},
\end{equation} 
with $C_{\mathrm{surr}}$ given by Eq.~\ref{model_error_covariance_matrix}.

The dimension-robust hierarchical MCMC sampling method implemented by \citet{han2023surrogate} is used to generate posterior samples of independent metaparameters and PCA latent variables (dimension-robust refers here to the fact that $\bm{\xi}$ can be of relatively high dimension). The method simulates a Markov chain to provide a sequence of samples that, in total, approximate the posterior distributions. The Metropolis-within-Gibbs treatment is applied~\citep{chen2018dimension}. This means that, at iteration $k+1$ of the Markov chain, we first generate a new sample of the PCA latent variables $\bm{\xi}^{'}$ by adding a random perturbation $\boldsymbol{\epsilon} \in \mathbb{R}^{n_d}$ to the latent variable $\bm{\xi}^{k}$ ($k$ denotes the previous iteration). This process can be represented as

\begin{equation} \label{xi}
\bm{\xi}^{'} = (1 - \beta^2) \bm{\xi}^{k} + \beta \boldsymbol{\epsilon},
\end{equation}

\noindent where $\beta$ is a coefficient that controls the magnitude of the update and $\boldsymbol{\epsilon} \in \mathbb{R}^{n_d}$ is a random vector with each component sampled independently from $\mathcal{N}$(0, 1). The new PCA latent variable sample $\bm{\xi}^{'}$ is then accepted with probability $\frac{p(\bm{\mathrm{d}}_{\mathrm{obs}} | {\boldsymbol{\uptheta}}_{\mathrm{meta}}^k, \bm{\xi}^{'})} {p(\bm{\mathrm{d}}_{\mathrm{obs}} | {\boldsymbol{\uptheta}}_{\mathrm{meta}}^k, \bm{\xi}^k)}$~\citep{hastings1970monte}, meaning it is accepted with certainty if the likelihood is larger for $\textbf{m}_f({\boldsymbol{\uptheta}}^k_{\mathrm{meta}}, \bm{\xi}^{'})$ than $\textbf{m}_f({\boldsymbol{\uptheta}}^k_{\mathrm{meta}}, \bm{\xi}^k)$. In this study, we set $\beta$ = 0.03 when considering both surface and subsurface observation data, resulting in an acceptance rate of about 20\%. This is within the expected range of 10\% to 40\% for MCMC-based methods~\citep{gelman1996efficient}.

Next, a new sample of the metaparameters, denoted ${\boldsymbol{\uptheta}}_{\mathrm{meta}}^{'}$, is proposed. This sample is generated by adding a random perturbation to the current sample ${\boldsymbol{\uptheta}}_{\mathrm{meta}}^{k}$. The proposal distribution is multivariate Gaussian and centered on the current sample, i.e.,  ${\boldsymbol{\uptheta}}_{\mathrm{meta}}^{'} \sim \mathcal{N}$(${\boldsymbol{\uptheta}}_{\mathrm{meta}}^{k}$, $C_{\uptheta}$), where $C_{\uptheta} \in \mathbb{R}^{7 \times 7}$ is the covariance matrix of the proposal distribution for the seven metaparameters. For metaparameter $i$ ($i = 1, \dots, 7$), we specify the standard deviation as $\sigma_i=(({{\uptheta}}_{\mathrm{meta}})_{i, \mathrm{max}}-({{\uptheta}}_{\mathrm{meta}})_{i, \mathrm{min}})/60$, where $({{\uptheta}}_{\mathrm{meta}})_{i, \mathrm{max}}$ and $({{\uptheta}}_{\mathrm{meta}})_{i, \mathrm{min}}$ are the maximum and minimum values of the prior range. The proposed metaparameter sample ${\boldsymbol{\uptheta}}_{\mathrm{meta}}^{'}$ is accepted based on an analogous criterion to that used for $\bm{\xi}^{'}$. This treatment results in acceptance rates that fall within the desired range. 

\subsection{History matching results for true model~1}
\label{`true' model 1}

True model~1 is a random geomodel realization with parameters sampled from the prior distributions. The metaparameters for this model are $\mu_{\log k}$ = 3.51, $\sigma_{\log k}$ = 1.94, $a_r$ = 0.04, $d$ = 0.021, $e$ = 0.062, $E_s$ = 9.22 GPa and $E_o$ = 25.54 GPa. As explained earlier, the true porosity for grid cells in the storage aquifer is related to the true log-permeability (shown in Fig.~\ref{logk_1}) through the expression $\phi = d \cdot \log_e k + e$. The mean and standard deviation of the true porosity field are $\mu_{\phi}$ = 0.14 and $\sigma_{\phi}$ = 0.04, respectively. 

Using both data types, the hierarchical MCMC history matching requires 95,213 iterations (surrogate model function evaluations) to achieve convergence in the metaparameter posterior distributions. A total of 21,028 sets of metaparameters and corresponding geomodel realizations are accepted during this procedure (i.e., about a 22\% acceptance rate). As noted earlier, the coupled flow-geomechanics simulations require 120~minutes per run (on 32 CPU cores), so this MCMC history matching would not be viable using high-fidelity simulations. Each surrogate model evaluation for the full solution requires about 0.15~second on a single Nvidia A100 GPU. The total serial elapsed time for hierarchical MCMC history matching (95,213 function evaluations) using the surrogate models is about 10.6~hours. This includes the time required to generate a geomodel realization (using PCA), execute MCMC, etc., at each iteration. Actual elapsed time can be longer depending on GPU availability and other cluster-related issues.

\begin{figure}[H]
\centering 
\subfloat[$\mu_{\mathrm{log}k}$ (surface data)]{\label{mean_logk_d}\includegraphics[width = 50mm]{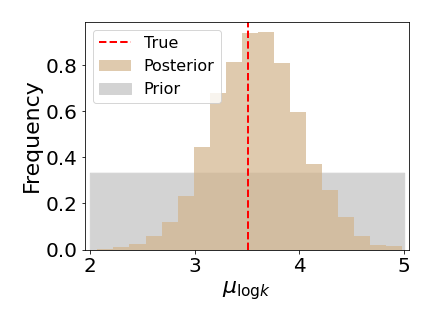}}
\hspace{5mm}
\subfloat[$\mu_{\mathrm{log}k}$ (subsurface data)]{\label{mean_logk_S_p}\includegraphics[width = 50mm]{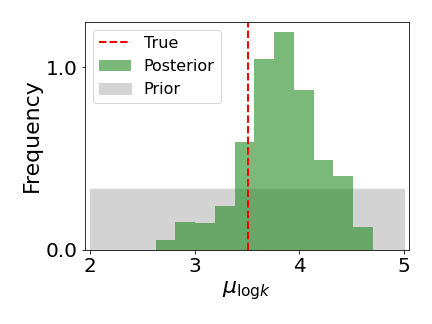}}
\hspace{4mm}
\subfloat[$\mu_{\mathrm{log}k}$ (both data types)]{\label{mean_logk_S_p_d}\includegraphics[width = 50mm]{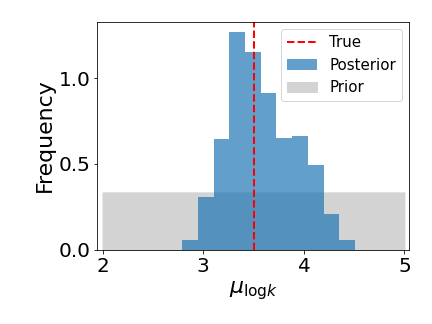}}\\
\subfloat[$\sigma_{\mathrm{log}k}$ (surface data)]{\label{std_logk_d}\includegraphics[width = 50mm]{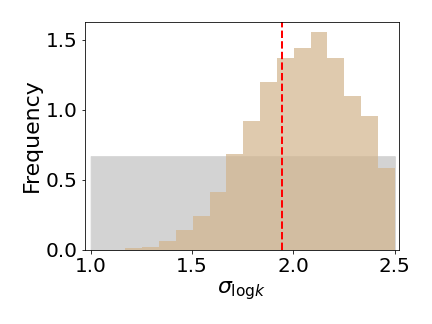}}
\hspace{5mm}
\subfloat[$\sigma_{\mathrm{log}k}$ (subsurface data)]{\label{std_logk_S_p}\includegraphics[width = 50mm]{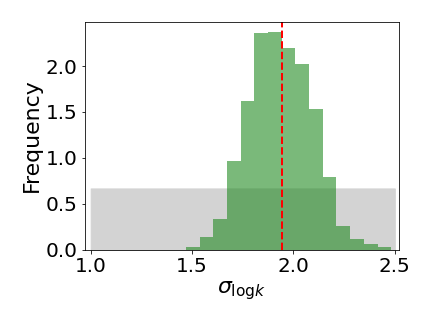}}
\hspace{5mm}
\subfloat[$\sigma_{\mathrm{log}k}$ (both data types)]{\label{std_logk_S_p_d}\includegraphics[width = 50mm]{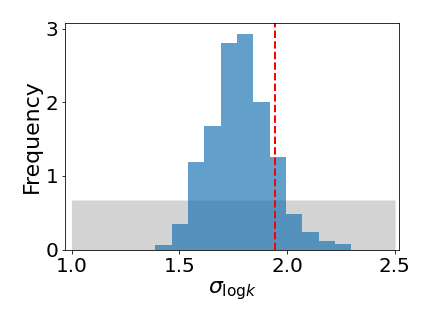}}\\
\subfloat[$\log_{10}(a_r)$ (surface data)]{\label{kvkh_d}\includegraphics[width = 50mm]{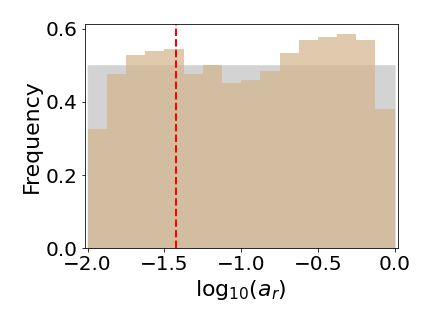}}
\hspace{5mm}
\subfloat[$\log_{10}(a_r)$ (subsurface data)]{\label{kvkh_S_p}\includegraphics[width = 50mm]{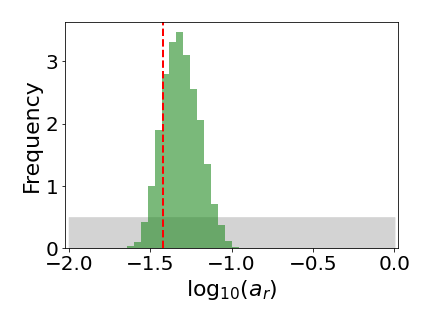}}
\hspace{5mm}
\subfloat[$\log_{10}(a_r)$ (both data types)]{\label{kvkh_S_p_d}\includegraphics[width = 50mm]{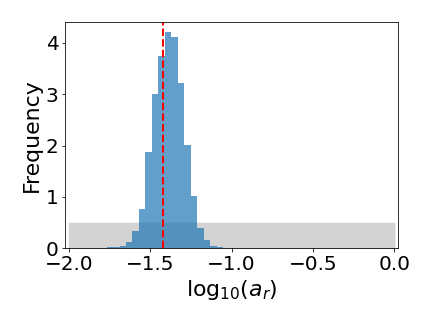}}\\
\caption{History matching results for the metaparameters $\mu_{\mathrm{log}k}$, $\sigma_{\mathrm{log}k}$, and $\log_{10}(a_r)$, for true model~1 using the total error covariance $C_{\mathrm{tot}}$ in the likelihood function (Eq.~\ref{model_error_covariance_matrix}). Gray regions represent prior distributions, brown histograms are posterior distributions using surface displacement data (first column), green histograms are posterior distributions using in-situ monitoring data (second column), blue histograms are posterior distributions using both data types (third column), and red vertical lines denote true values.}
\label{meta_true_1}
\end{figure}

\begin{figure}[H]
\centering  
\subfloat[$\mu_{\phi}$ (surface data)]{\label{mean_phi_d}\includegraphics[width = 50mm]{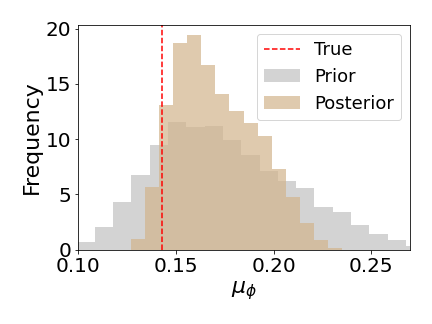}}
\hspace{5mm}
\subfloat[$\mu_{\phi}$ (subsurface data)]{\label{mean_phi_S_p}\includegraphics[width = 50mm]{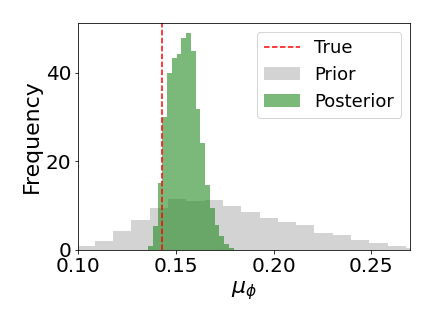}}
\hspace{5mm}
\subfloat[$\mu_{\phi}$ (both data types)]{\label{mean_phi_S_p_d}\includegraphics[width = 50mm]{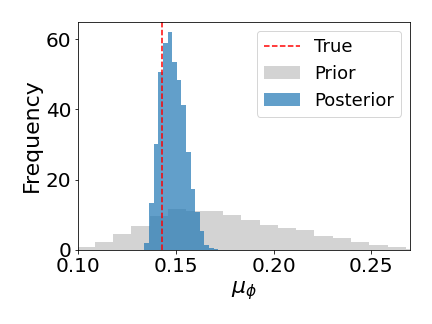}}\\
\subfloat[$\sigma_{\phi}$ (surface data)]{\label{std_phi_d}\includegraphics[width = 50mm]{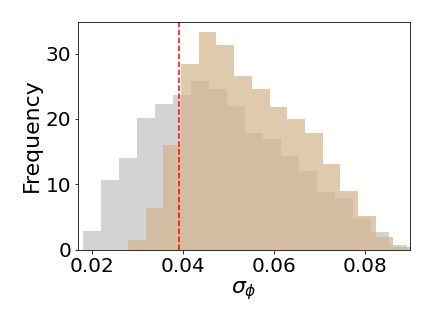}}
\hspace{5mm}
\subfloat[$\sigma_{\phi}$ (subsurface data)]{\label{std_phi_S_p}\includegraphics[width = 50mm]{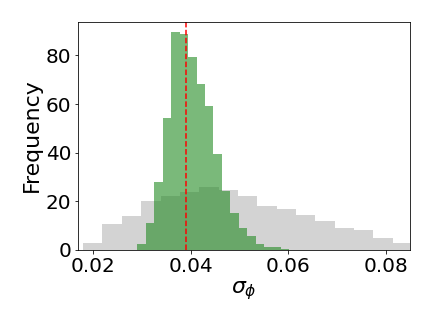}}
\hspace{5mm}
\subfloat[$\sigma_{\phi}$ (both data types)]{\label{std_phi_S_p_d}\includegraphics[width = 50mm]{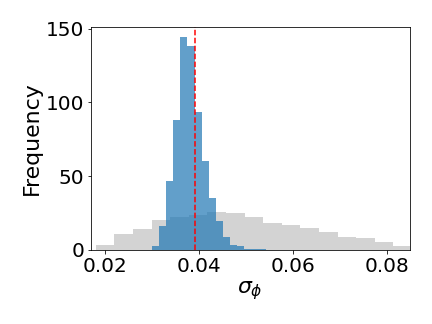}}\\
\subfloat[$E_s$ (surface data)]{\label{E_s_d}\includegraphics[width = 50mm]{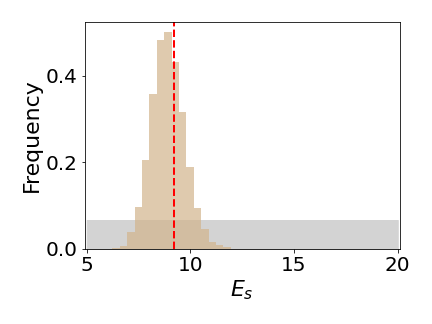}}
\hspace{5mm}
\subfloat[$E_s$ (subsurface data)]{\label{E_s_S_p}\includegraphics[width = 50mm]{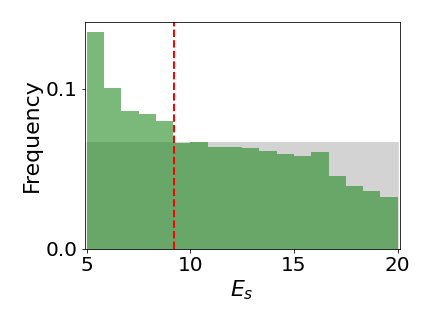}}
\hspace{5mm}
\subfloat[$E_s$ (both data types)]{\label{E_s_S_p_d}\includegraphics[width = 50mm]{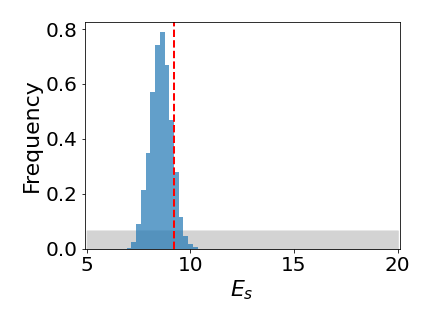}}\\
\subfloat[$E_o$ (surface data)]{\label{E_o_d}\includegraphics[width = 50mm]{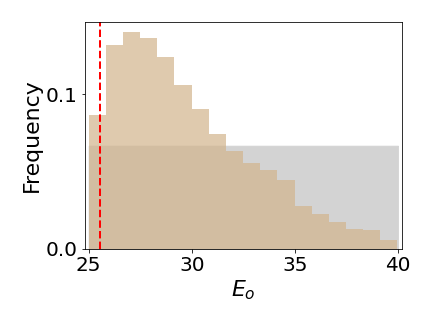}}
\hspace{5mm}
\subfloat[$E_o$ (subsurface data)]{\label{E_o_S_p}\includegraphics[width = 50mm]{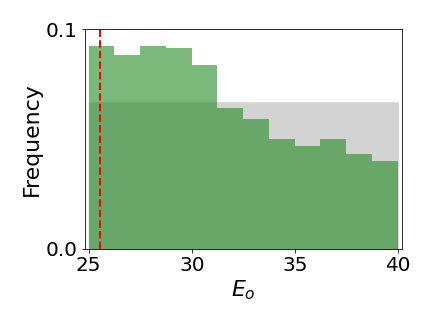}}
\hspace{5mm}
\subfloat[$E_o$ (both data types)]{\label{E_o_S_p_d}\includegraphics[width = 50mm]{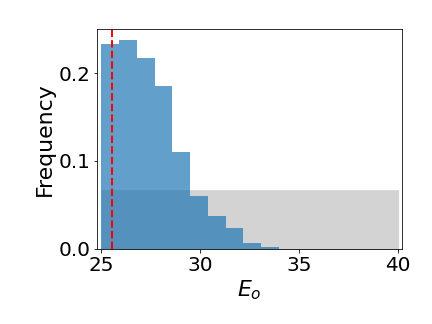}}
\caption{History matching results for $\mu_{\phi}$ and $\sigma_{\phi}$, and metaparameters $E_s$ and $E_o$ (Young's moduli in the storage aquifer and overburden), for true model~1 using the total error covariance $C_{\mathrm{tot}}$ in the likelihood function (Eq.~\ref{model_error_covariance_matrix}). Gray regions represent prior distributions, brown histograms are posterior distributions using surface displacement data (first column), green histograms are posterior distributions using in-situ monitoring data (second column), blue histograms are posterior distributions using both data types (third column), and red vertical lines denote true values.}
\label{meta_1_true_1}
\end{figure}

History matching results for the metaparameters for this case, accounting for both the measurement errors and surrogate model errors as described above, are presented in Figs.~\ref{meta_true_1} and~\ref{meta_1_true_1}. The posterior distribution of the metaparameters is shown for three different observation data types and combinations: in-situ monitoring-well data only (this includes saturation in O1--O4 and pressure in O5), surface displacement data only (at 81 surface locations), and both data types together. It is important to note that different $C_{\mathrm{surr}}$ matrices are constructed from the test-set data (using Eq.~\ref{model_error_covariance_matrix}) for each of these three approaches. The gray regions in each plot indicate the prior distributions, the brown histograms display posterior distributions using only surface displacement data, the green histograms are posterior distributions with only in-situ monitoring-well data, the blue histograms represent the posterior distributions using both data types, and the red vertical dashed lines indicate the true parameter values. Note that the computed mean ($\mu_{\phi}$) and standard deviation ($\sigma_{\phi}$) of the porosity field are displayed in Fig.~\ref{meta_1_true_1}, rather than the metaparameters $d$ and $e$. This is because, consistent with \citet{han2023surrogate}, the measured data are more informative for these quantities than for $d$ and $e$ individually.

The results in Figs.~\ref{meta_true_1} and~\ref{meta_1_true_1} clearly display the impact of the various data types and combinations on posterior metaparameter uncertainty. The use of both data types provides substantial uncertainty reduction for all parameters (with the possible of exception of $E_o$, Young's modulus in the overburden, shown in Fig.~\ref{meta_1_true_1}l). The use of surface data only is somewhat informative for $\mu_{\log k}$, $\sigma_{\log k}$ and $E_s$ (Fig.~\ref{meta_true_1}a and d, Fig.~\ref{meta_1_true_1}g), but it provides less information for other metaparameters. Subsurface data by itself inform all quantities to some degree, with the exception of $E_s$ and $E_o$ (Fig.~\ref{meta_1_true_1}h and k). Importantly, the posterior distributions are, for all quantities, narrower when both data types are used. This behavior would be expected intuitively, but it may not always be observed if model error is not correctly included in the history matching procedure. Note finally that, for all data types and combinations, the true values fall within the posterior distributions.

\begin{figure}[H]
\centering   
\subfloat[Prior 1]{\label{Prior_logk_1}\includegraphics[width=31mm]{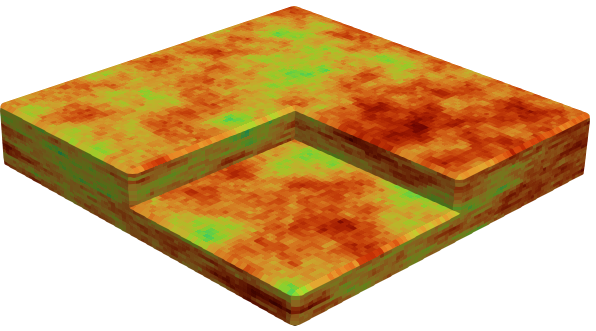}}
\hspace{1mm}
\subfloat[Prior 2]{\label{Prior_logk_2}\includegraphics[width=31mm]{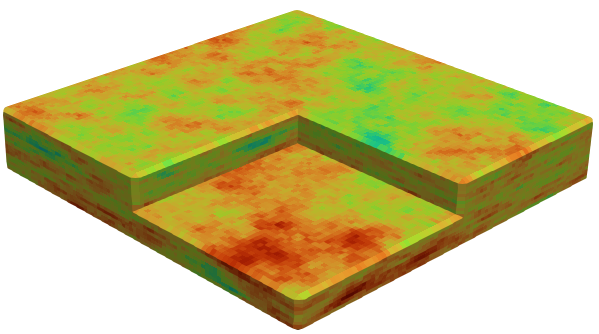}}
\hspace{1mm}
\subfloat[Prior 3]{\label{Prior_logk_3}\includegraphics[width=31mm]{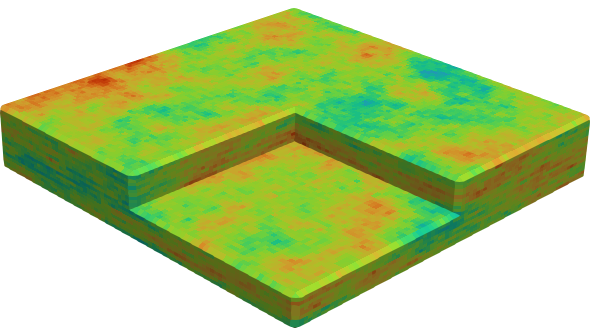}}
\hspace{1mm}
\subfloat[Prior 4]{\label{Prior_logk_4}\includegraphics[width=31mm]{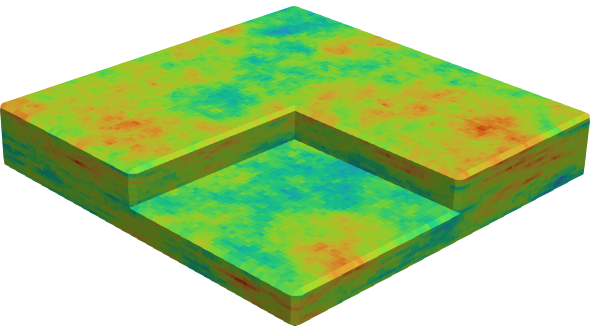}}
\hspace{1mm}
\subfloat[Prior 5]{\label{Prior_logk_5}\includegraphics[width=31mm]{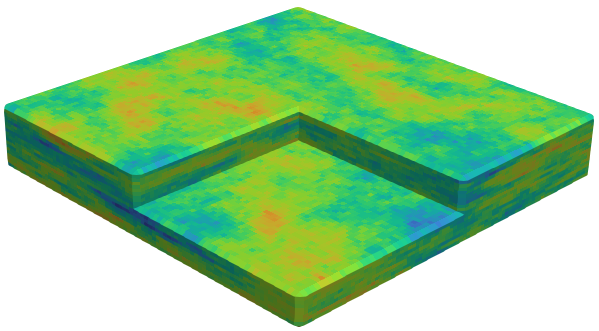}}
\includegraphics[width=5.5mm]{Figure/Surrogate/logk_Scale.PNG}\\[1ex]
\subfloat[Posterior 1]{\label{Posterior_logk_1}\includegraphics[width=31mm]{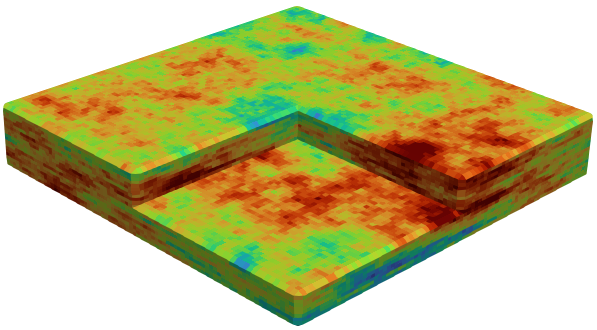}}
\hspace{1mm}
\subfloat[Posterior 2]{\label{Posterior_logk_2}\includegraphics[width=31mm]{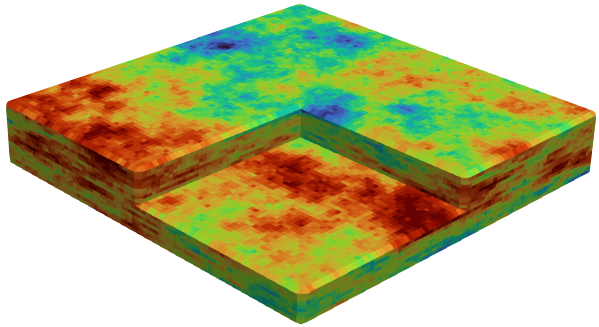}}
\hspace{1mm}
\subfloat[Posterior 3]{\label{Posterior_logk_3}\includegraphics[width=31mm]{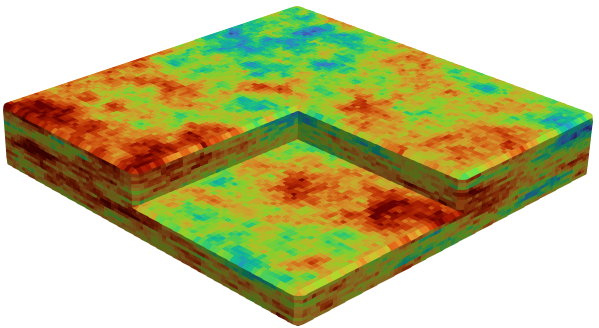}}
\hspace{1mm}
\subfloat[Posterior 4]{\label{Posterior_logk_4}\includegraphics[width=31mm]{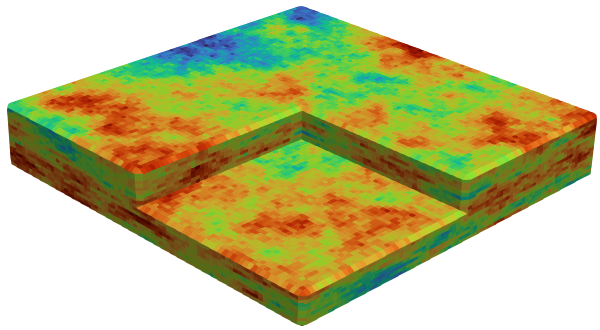}}
\hspace{1mm}
\subfloat[Posterior 5]{\label{Posterior_logk_5}\includegraphics[width=31mm]{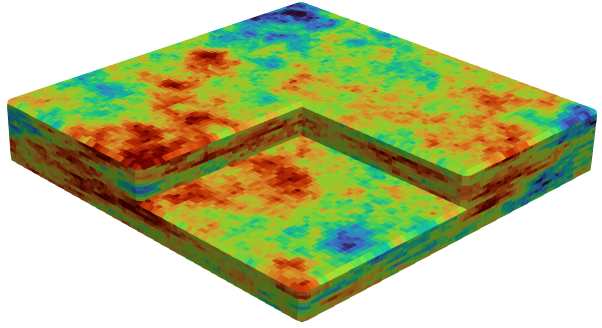}}
\includegraphics[width=5.5mm]{Figure/Surrogate/logk_Scale.PNG}
\caption{Representative log-permeability fields for the prior geomodels (upper row) and posterior geomodels (lower row), for true model~1. True log-permeability field for true model~1 is shown in Fig.~\ref{logk_1}.}
\label{Prior_Posterior:logk}
\end{figure}

In addition to the metaparameters, the hierarchical MCMC procedure also provides posterior geological realizations. We identify `representative' realizations by applying a k-means clustering method and then selecting the medoid for each cluster. Five such log-permeability fields for prior geomodels (cluster medoids from the test set), and five fields for posterior geomodels (cluster medoids for 500 randomly selected posterior realizations), are shown in Fig.~\ref{Prior_Posterior:logk}. The prior log-permeability fields have clearly different means and display a high level of variability. The variation between realizations is reduced considerably in the posterior fields. A reasonable degree of visual consistency with the true log-permeability field (shown in Fig.~\ref{logk_1}) is evident, particularly in the Posterior~1, 2 and 3 realizations in Fig.~\ref{Prior_Posterior:logk}. 

\begin{figure}[H]
\centering
\subfloat[Saturation at O1 in layer 1]{\label{s_O1}\includegraphics[width = 78mm]{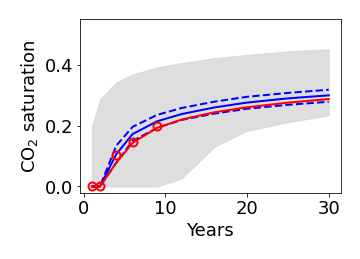}}
\hspace{2mm}
\subfloat[Saturation at O2 in layer 6]{\label{s_O2}\includegraphics[width = 78mm]{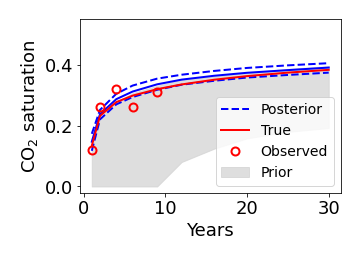}}
\\[1ex]
\subfloat[Saturation at O3 in layer 12]{\label{s_O3}\includegraphics[width = 78mm]{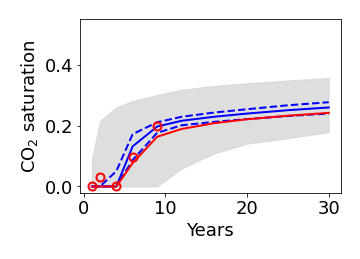}}
\hspace{2mm}
\subfloat[Saturation at O4 in layer 19]{\label{s_O4}\includegraphics[width = 78mm]{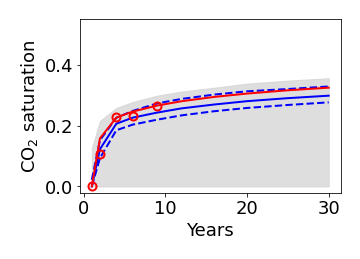}}
\\[1ex]
\caption{History matching results for saturation in different layers at monitoring wells~O1--O4 for true model~1 using both data types. Gray regions represent the prior P$_{10}$--P$_{90}$ range, red circles and red curves denote observed and true data, and blue curves show the posterior P$_{10}$, P$_{50}$ and P$_{90}$ predictions. Legend in (b) applies to all subplots.}
\label{Posterior_Sw}
\end{figure}

We now present detailed posterior results for saturation and surface displacement. History matching results (using both data types) for saturation in different layers at the four monitoring wells are shown in Fig.~\ref{Posterior_Sw}. The gray regions indicate the prior P$_{10}$--P$_{90}$ ranges, the red curves represent the true saturation from the coupled flow-geomechanics simulation, and the red circles are the observed saturation data. These points include measurement error, which is why they deviate from the red curves. The blue solid curves are P$_{50}$ posterior results, and the blue dashed curves are P$_{10}$ (lower) and P$_{90}$ (upper) posterior results. The priors show a large amount of uncertainty, which is reduced considerably in the posterior results. During both the history matching period (first 9~years) and the prediction period (up to 30~years), the true data and true response are largely captured within the posterior P$_{10}$--P$_{90}$ ranges.

\begin{figure}[H]
\centering   
\subfloat[Prior 1]{\label{prior_s_1}\includegraphics[width=30mm]{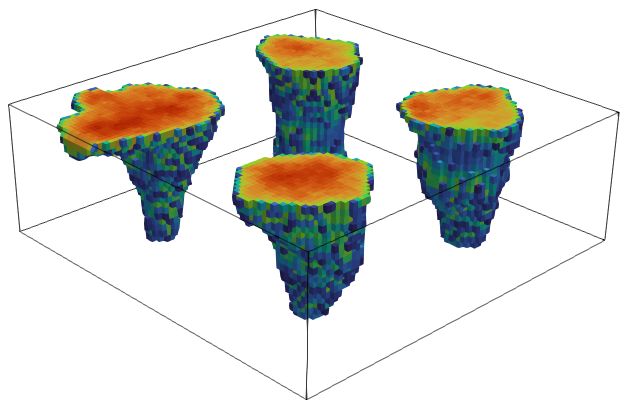}}
\hspace{1.5mm}
\subfloat[Prior 2]{\label{prior_s_2}\includegraphics[width=30mm]{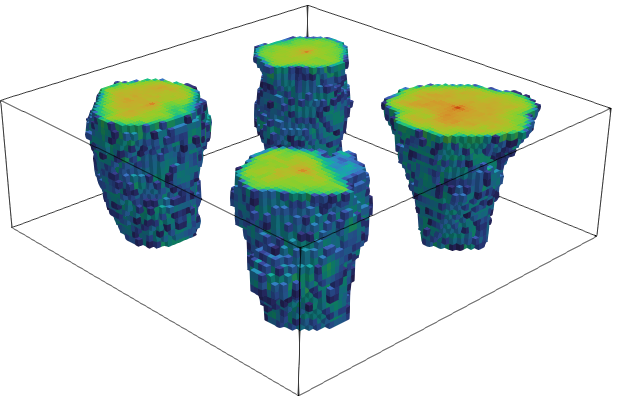}}
\hspace{1.5mm}
\subfloat[Prior 3]{\label{prior_s_3}\includegraphics[width=30mm]{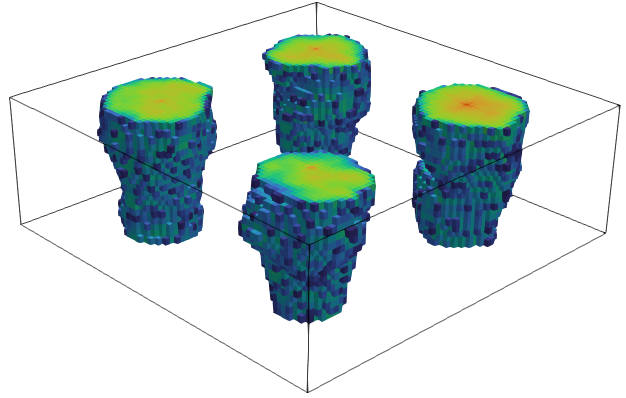}}
\hspace{1.5mm}
\subfloat[Prior 4]{\label{prior_s_4}\includegraphics[width=30mm]{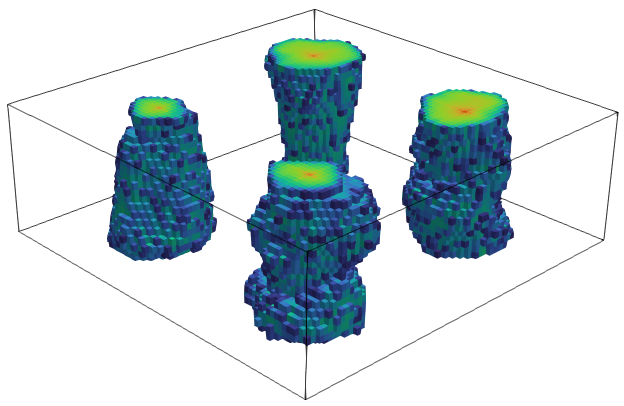}}
\hspace{1.5mm}
\subfloat[Prior 5]{\label{prior_s_5}\includegraphics[width=30mm]{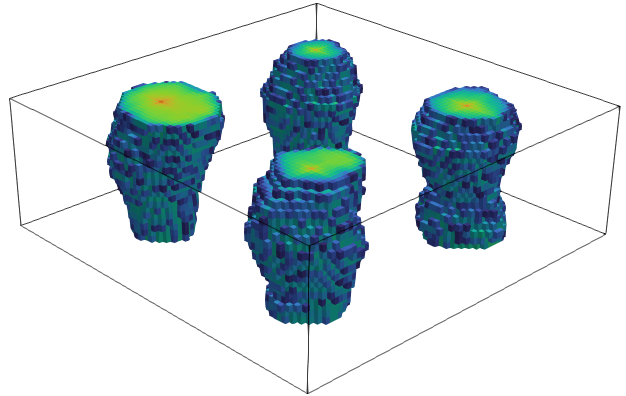}}
\includegraphics[width=6.5mm]{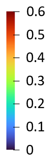}\\[1ex]
\subfloat[Posterior 1]{\label{posterior_s_1}\includegraphics[width=30mm]{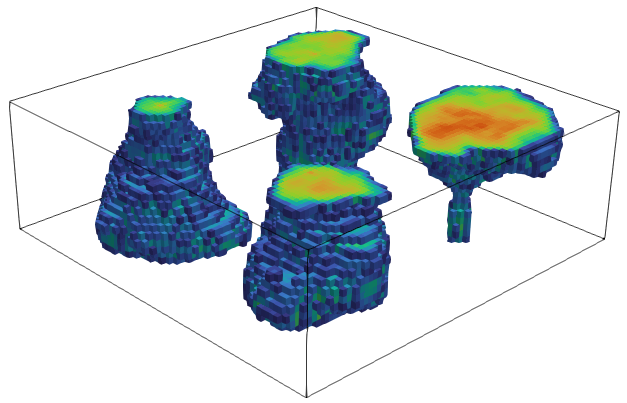}}
\hspace{1.5mm}
\subfloat[Posterior 2]{\label{posterior_s_2}\includegraphics[width=30mm]{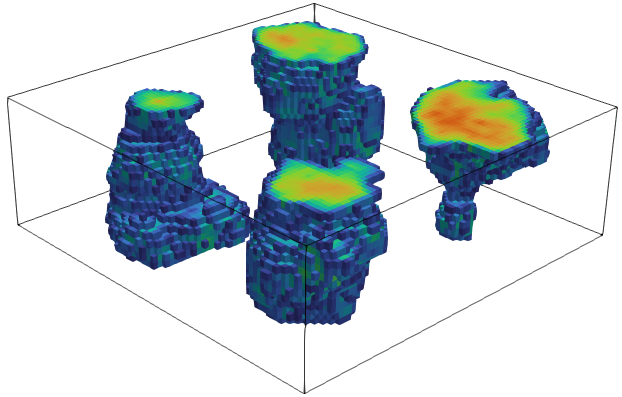}}
\hspace{1.5mm}
\subfloat[Posterior 3]{\label{posterior_s_3}\includegraphics[width=30mm]{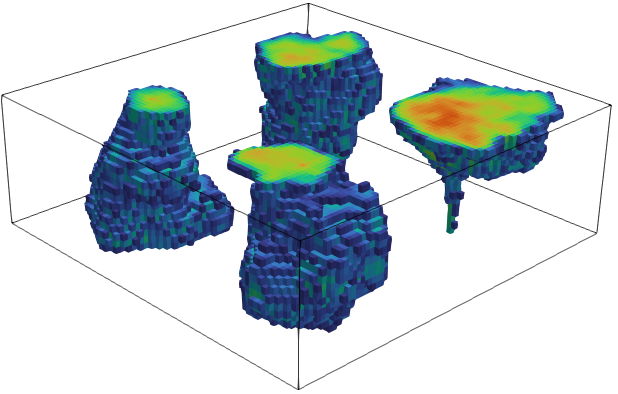}}
\hspace{1.5mm}
\subfloat[Posterior 4]{\label{posterior_s_4}\includegraphics[width=30mm]{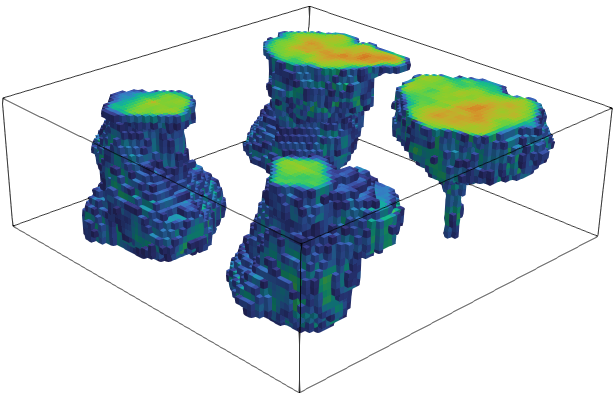}}
\hspace{1.5mm}
\subfloat[Posterior 5]{\label{posterior_s_5}\includegraphics[width=30mm]{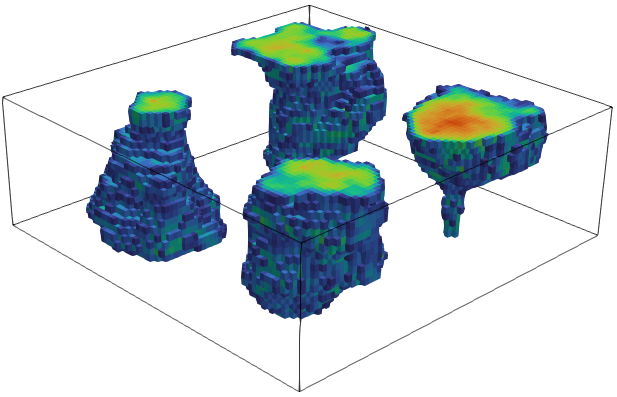}}
\includegraphics[width=6.5mm]{Figure/Surrogate/Sw_Scale.PNG}
\caption{Representative CO$_2$ saturation fields at 30~years, from prior geomodels (upper row) and posterior geomodels (lower row), for true model~1 using both data types. True saturation field at 30~years for true model~1 is shown in Fig.~\ref{sim_1}.}
\label{Prior_Posterior:Saturation}
\end{figure}

Representative saturation fields for prior and posterior geomodels, at the end of the 30-year injection period, are shown in Fig.~\ref{Prior_Posterior:Saturation}. These fields are selected using the procedure described above for the identification of representative log-permeability fields. The true saturation field is shown in Fig.~\ref{sim_1}. The prior saturation fields are quite regular and symmetric. The posterior plumes, by contrast, vary from injector to injector, and all contain the `thin-stemmed' plume on the right and the somewhat bulbous plume in front. It is noteworthy that data from five in-situ monitoring wells, along with surface displacement measurements, can provide this relatively high degree of characterization for plumes with such complicated shapes.

\begin{figure}[H]
\centering
\subfloat[Surface displacement above I1]{\label{d_1}\includegraphics[width = 78mm]{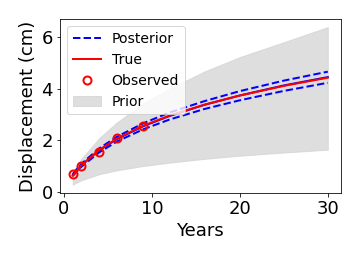}}
\hspace{2mm}
\subfloat[Surface displacement above I2]{\label{d_2}\includegraphics[width = 78mm]{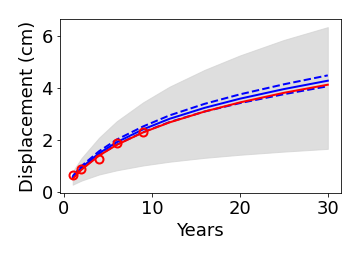}}
\\[1ex]
\subfloat[Surface displacement above I3]{\label{d_3}\includegraphics[width = 78mm]{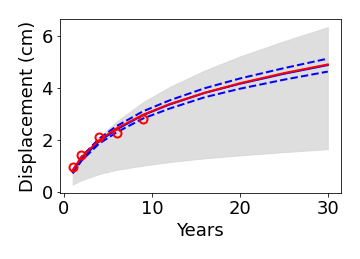}}
\hspace{2mm}
\subfloat[Surface displacement above I4]{\label{d_4}\includegraphics[width = 78mm]{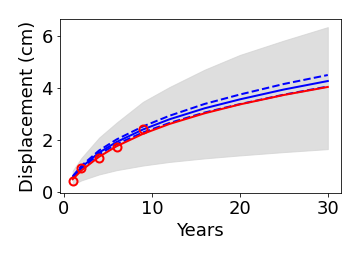}}
\\[1ex]
\caption{History matching results for surface displacement at observation locations above the injectors, for true model~1 using both data types. Gray regions represent the prior P$_{10}$--P$_{90}$ range, red circles and red curves denote observed and true data, and blue curves show the posterior P$_{10}$, P$_{50}$ and P$_{90}$ predictions. Legend in (a) applies to all subplots.}
\label{Posterior_d}
\end{figure}

\begin{figure}[H]
\centering   
\subfloat[Prior 1]{\label{prior_d_1}\includegraphics[width=32mm]{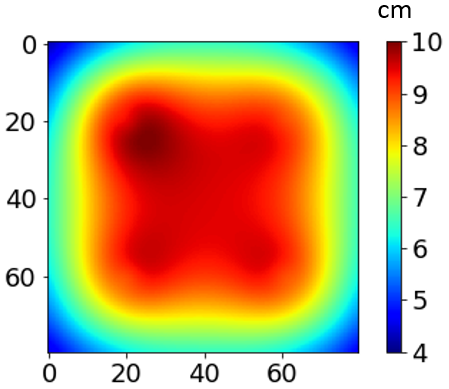}}
\hspace{2mm}
\subfloat[Prior 2]{\label{prior_d_2}\includegraphics[width=31mm]{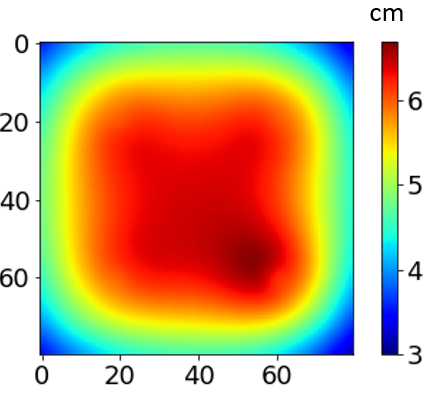}}
\hspace{2mm}
\subfloat[Prior 3]{\label{prior_d_3}\includegraphics[width=31mm]{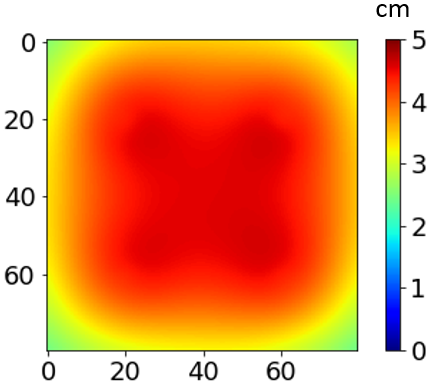}}
\hspace{2mm}
\subfloat[Prior 4]{\label{prior_d_4}\includegraphics[width=31mm]{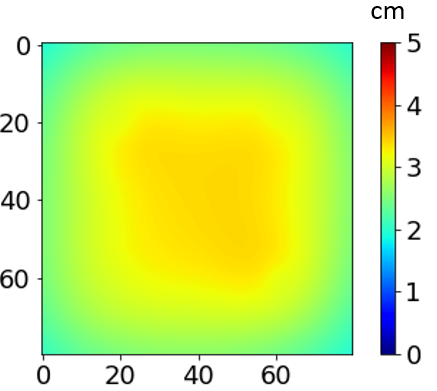}}
\hspace{2mm}
\subfloat[Prior 5]{\label{prior_d_5}\includegraphics[width=31mm]{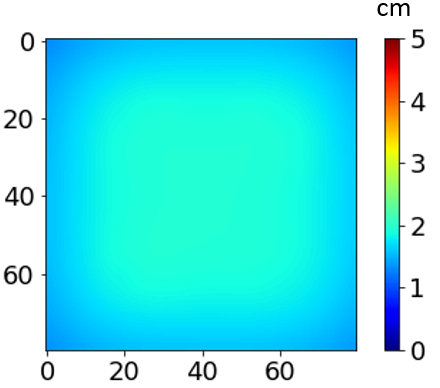}}\\[1ex]
\subfloat[Posterior 1]{\label{posterior_d_1}\includegraphics[width=31mm]{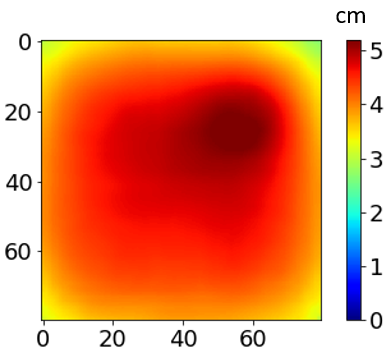}}
\hspace{2mm}
\subfloat[Posterior 2]{\label{posterior_d_2}\includegraphics[width=31mm]{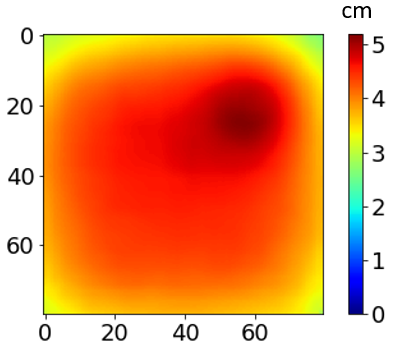}}
\hspace{2mm}
\subfloat[Posterior 3]{\label{posterior_d_3}\includegraphics[width=31mm]{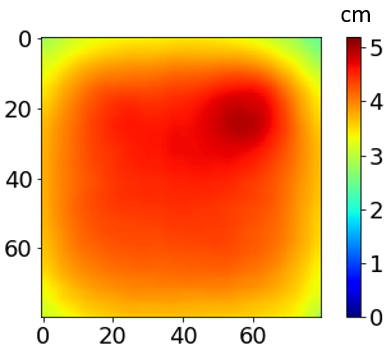}}
\hspace{2mm}
\subfloat[Posterior 4]{\label{posterior_d_4}\includegraphics[width=31mm]{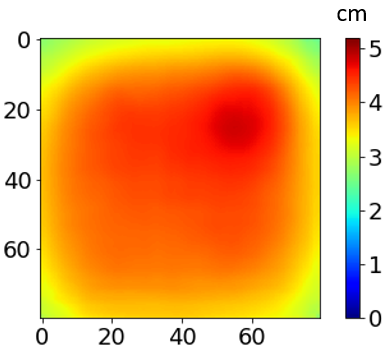}}
\hspace{2mm}
\subfloat[Posterior 5]{\label{posterior_d_5}\includegraphics[width=31mm]{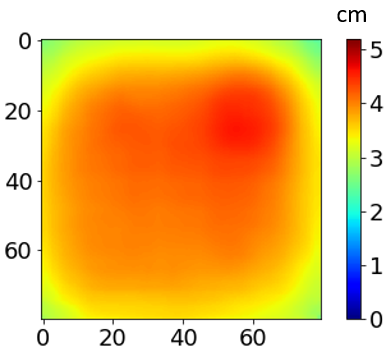}}
\caption{Representative surface displacement fields at 30~years, from prior geomodels (upper row) and posterior geomodels (lower row), for true model~1 using both data types. True surface displacement field at 30~years for true model~1 is shown (with a different colorbar scale) in Fig.~\ref{sim_d_1}.}
\label{Prior_Posterior:Displacement}
\end{figure}

Results for surface displacement up to 30~years, at locations directly above the injection wells, are shown in Fig.~\ref{Posterior_d}. We observe very substantial uncertainty reduction for these quantities, with the true solution largely captured within the posterior P$_{10}$--P$_{90}$ range. Five representative prior and posterior surface displacement fields at 30~years (identified as described above) are presented in Fig.~\ref{Prior_Posterior:Displacement}. The prior fields display a wide range of responses -- a somewhat uniform surface displacement of $\sim$2~cm for Prior~5, and up to 10~cm of displacement in some regions in Prior~1. The true model results are shown in Fig.~\ref{sim_d_1} (note the different colorbar scale). The posterior fields in Fig.~\ref{Posterior_d}f-j resemble the true results, and capture the large displacement in the upper right of the model. The posterior results display some variability at other surface locations.

Note that the metaparameter priors are all uniformly distributed in this study. We considered both uniformly-distributed and Gaussian-distributed priors in the hierarchical MCMC-based history matching method in our earlier (flow-only) study. In that case similar performance was observed with both distributions.

History matching results for true model~1 with surrogate model error neglected are presented in SI. There we observe more narrow (and somewhat nonphysical for some metaparameters) posterior distributions, indicating the importance of including surrogate model error in history matching for this case. We also present hierarchical MCMC-based history matching for true model~2, which is assigned correlation lengths that differ from those in all prior models. History matching results for this case are still acceptable, though some parameters are less accurately estimated. This is presumably because the MCMC procedure must `compensate' for the fact that the true model is outside the prior.

\section{Concluding Remarks} 
\label{Conclusions}
In this study, we developed a new deep learning surrogate modeling framework that accepts an input geomodel and provides predictions for saturation, pressure and surface displacement in a carbon storage operation. These predicted quantities are then used for MCMC-based history matching. An important feature of our approach is that relatively few coupled flow-geomechanics simulation runs are required for network training. A large number of flow-only simulations, with an effective rock compressibility treatment, are used instead. This has a large impact on training time since the fully coupled models require $15\times$ as much computation as the flow-only runs. A new network architecture for surface displacement, which accepts pressure and saturation surrogate model predictions as inputs, was developed for use in this workflow. The surrogate models were trained over a broad set of prior geomodels that are characterized by seven uncertain metaparameters, including mean and standard deviation of log-permeability, permeability anisotropy ratio, and Young’s moduli in the storage aquifer and overburden.

We considered a large-scale 30-year storage project involving four CO$_2$ injectors, with a total injection rate of 4~Mt/year. We assessed test-set error (involving 500 new geomodel realizations) for several different training strategies. Using 4000 flow-only runs and 400 fully coupled runs for training, we achieved median relative errors for saturation, pressure and surface displacement of 3.9\%, 0.8\% and 2.6\%, respectively. Surface displacement results of nearly this accuracy were achieved using 200 fully coupled training runs (along with 4000 flow-only runs). The time required to perform 4000 flow-only and 200 fully coupled training runs is about an order of magnitude less than that needed to simulate 4000 fully coupled models, so our framework offers significant benefit in this regard.

The surrogate models were then incorporated into a hierarchical MCMC history matching workflow. A new treatment to account for surrogate error, involving the full model error covariance matrix, was introduced. This important issue does not appear to have been considered in previous surrogate-based history matching procedures. For a synthetic true model, we compared posterior predictions for the metaparameters using three different data types/combinations –- in-situ monitoring-well data only, surface displacement data only, and both data types together. The use of both data types was shown to provide the most uncertainty reduction for all parameters, as would be expected. The posterior saturation plumes and surface displacement fields were also seen to be in reasonable correspondence with the true results. In SI, we show that our new treatment for surrogate model error in history matching leads to improved posterior results. In addition, history matching results for a synthetic true model that is outside the prior are also shown to be acceptable.

There are many useful directions for future research in this area. It will be of interest to consider more complicated geomodels, such as those containing extensive fractures or faults. This will introduce additional uncertain parameters into the workflow, and may necessitate extension of the hierarchical MCMC procedure. It may also lead to reduced accuracy in the pressure and saturation fields from flow-only (effective rock compressibility) simulations. Thus we may need to implement correction procedures for the flow variables, which could involve additional networks. The ability to handle 4D seismic data, along with stress and strain data from in-situ monitoring-wells, should also be incorporated into the overall framework. Finally, it will be useful to extend and test our treatments on practical cases.

\section*{CRediT authorship contribution statement}
\textbf{Yifu Han}: Conceptualization, Methodology, Software, Visualization, Formal analysis, Writing -- original draft. \textbf{Fran\c cois P. Hamon}: Conceptualization, GEOS software, Writing -- original draft. \textbf{Louis J. Durlofsky}: Supervision, Conceptualization, Resources, Formal analysis, Writing -- review \& editing.

\section*{Declaration of competing interest}
The authors declare that they have no known competing financial interests or personal relationships that could have appeared to influence the work reported in this paper.

\section*{Data availability}
The code used in this study will be made available on github when this paper is published. Please contact Yifu Han (yifu@stanford.edu) for earlier access.

\section*{Acknowledgements} 
We thank the Stanford Center for Carbon Storage, Stanford Smart Fields Consortium, and TotalEnergies (through the FC-MAELSTROM project) for funding. We are grateful to the SDSS
Center for Computation for HPC resources, and to developers at Lawrence Livermore National Laboratory, Stanford University, and TotalEnergies for assistance with GEOS.

\section*{Supplementary Information} 
\section*{SI 1. SI overview}
\renewcommand{\thesection}{SI \arabic{section}} 
In this SI, we present additional history matching results for two cases -- true model~1 with surrogate model error neglected, and true model~2, which is characterized by a correlation structure that is different than the prior. All other specifications are the same as those used in the main paper.

\section*{SI 2. Impact of model error on posterior metaparameters}
\renewcommand{\thesection}{SI \arabic{section}} 
\label{model_error_impact}
Model error was included in all results in the main paper. We now present results in which this effect is neglected, which will allow for an assessment of its impact. Model error is ignored by setting $C_{\mathrm{surr}}=0$ in Eq.~25 in the main paper. The hierarchical MCMC procedure is otherwise identical to that applied earlier for true model~1. 

\begin{figure}[H]
\centering 
\subfloat[$\mu_{\mathrm{log}k}$ (both data types)]{\label{mean_logk_CD}\includegraphics[width = 50mm]{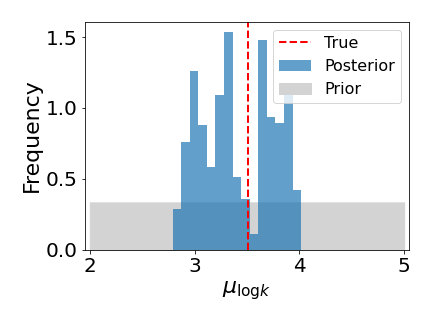}}
\hspace{5mm}
\subfloat[$\sigma_{\mathrm{log}k}$ (both data types)]{\label{std_logk_CD}\includegraphics[width = 50mm]{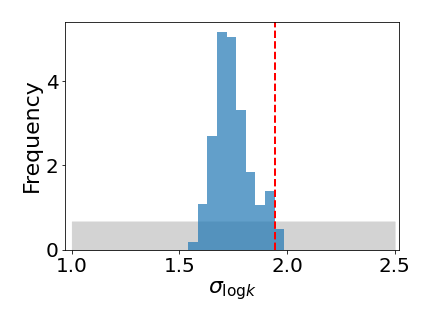}}
\hspace{5mm}
\subfloat[$\log_{10}(a_r)$ (both data types)]{\label{kvkh_CD}\includegraphics[width = 50mm]{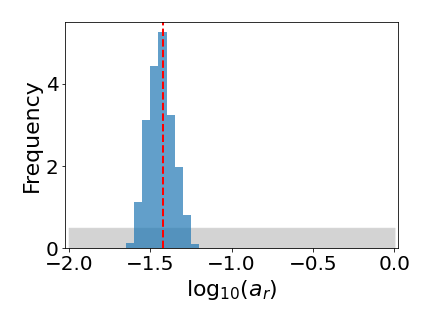}}\\
\subfloat[$\mu_{\phi}$ (both data types)]{\label{mean_phi_CD}\includegraphics[width = 50mm]{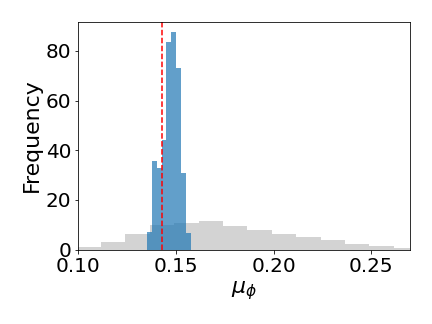}}
\hspace{10mm}
\subfloat[$\sigma_{\phi}$ (both data types)]{\label{std_phi_CD}\includegraphics[width = 50mm]{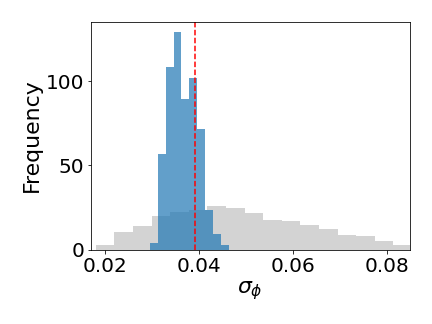}}\\
\subfloat[$E_s$ (both data types)]{\label{E_s_CD}\includegraphics[width = 50mm]{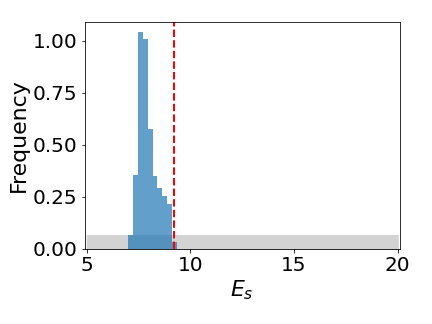}}
\hspace{10mm}
\subfloat[$E_o$ (both data types)]{\label{E_o_CD}\includegraphics[width = 50mm]{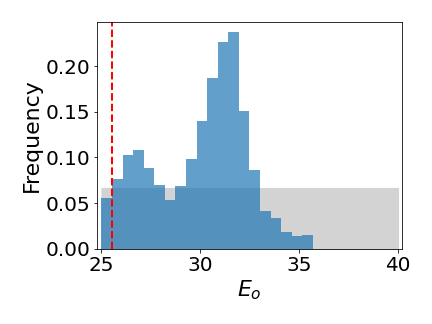}}
\caption{History matching results for the metaparameters and porosity parameters for true model~1. Model error contribution is neglected in these results ($C_{\mathrm{surr}}=0$ in Eq.~25 in the main paper). Gray regions represent prior distributions, blue histograms are posterior distributions using both data types, and red vertical lines denote true values. Legend in (a) applies to all subplots.}
\label{meta_true_1_CD}
\end{figure}

Posterior results for the metaparameters are shown in Fig.~\ref{meta_true_1_CD}. These results, for which both data types are used, should be compared to results in the third column in Figs.~13 and~14 in the main paper. The first observation is that the posterior uncertainty ranges in Fig.~\ref{meta_true_1_CD} are generally less than those in Figs.~13 and~14. This is consistent with the observed data being afforded more precision in the present case, i.e., with $C_{\mathrm{surr}}=0$. Another observation is that some of the posterior distributions in Fig.~\ref{meta_true_1_CD} are jagged or bimodal (e.g., $\mu_{\log k}$, $E_o$), in contrast to the results in Figs.~13 and~14. 

In an overall sense, the results in the main paper, in which $C_{\mathrm{surr}}$ is included, appear to be more physically realistic. In those results, consistent with expectations, we do not observe bimodal or highly discontinuous histograms. Related observations on the importance of model error were made by \citet{rammay2019quantification}.

\section*{SI 3. History matching results for true model~2}
\renewcommand{\thesection}{SI \arabic{section}} 
\label{true model_2}
We now present history matching results for a second true model. The horizontal and vertical correlation lengths in the storage aquifer for true model~2 differ from the (fixed) values used for the prior models, so this true model is outside the prior range and thus represents a more challenging case. For this model, the horizontal correlation length is set to 64~cells (9600~m, which is double the $l_h$ of 4800~m for geomodels in the prior and in the training set), and the vertical correlation length is set to 4~cells (20~m, which is half the $l_v$ of 40~m for geomodels in the prior and the training set). The metaparameter values for true model~2 are $\mu_{\log k}$ = 2.31, $\sigma_{\log k}$ = 2.05, $a_r$ = 0.05, $d$ = 0.036, $e$ = 0.076, $E_s$ = 7.63~GPa and $E_o$ = 28.3~GPa. The log-permeability field of the storage aquifer for this model is shown in Fig.~\ref{logk_True_2}. The mean and standard deviation of the porosity field, constructed based on the log-permeability field, $d$ and $e$, are $\mu_{\phi}=0.15$ and $\sigma_{\phi}=0.08$.

In the MCMC procedure, the horizontal and vertical correlation lengths are not updated, so the history matching must adjust other parameters to account for these unmodeled effects. Model error (as represented in $C_{\mathrm{surr}}$) is included in this case, and both data types are used. A total of 64,072 MCMC iterations are required to achieve convergence in the posterior distributions. A total of 17,232 sets of metaparameters and corresponding geomodel realizations are accepted (acceptance rate of about 27\%).

\begin{figure}[H]
\centering 
\subfloat[$\mu_{\mathrm{log}k}$ (both data types)]{\label{mean_logk}\includegraphics[width = 50mm]{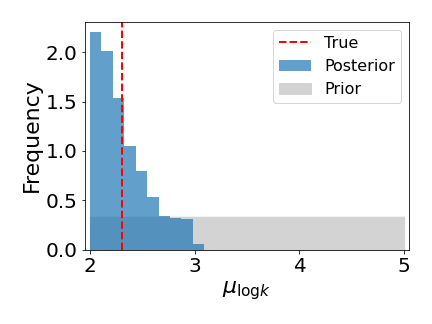}}
\hspace{5mm}
\subfloat[$\sigma_{\mathrm{log}k}$ (both data types)]{\label{std_logk}\includegraphics[width = 50mm]{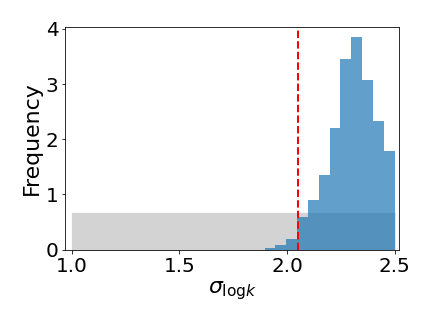}}
\hspace{5mm}
\subfloat[$\log_{10}(a_r)$ (both data types)]{\label{kvkh}\includegraphics[width = 50mm]{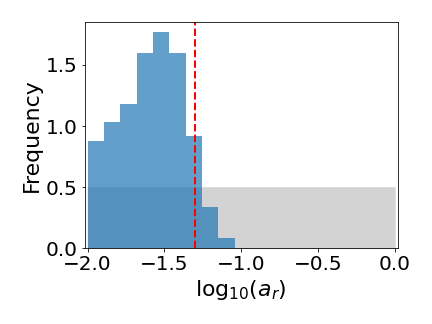}}\\
\subfloat[$\mu_{\phi}$ (both data types)]{\label{mean_phi}\includegraphics[width = 50mm]{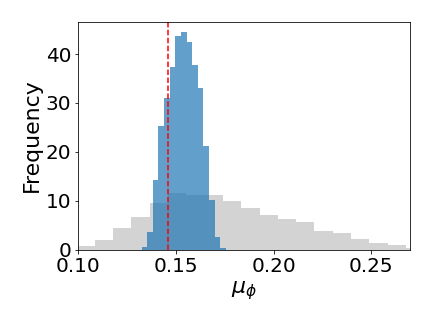}}
\hspace{10mm}
\subfloat[$\sigma_{\phi}$ (both data types)]{\label{std_phi}\includegraphics[width = 50mm]{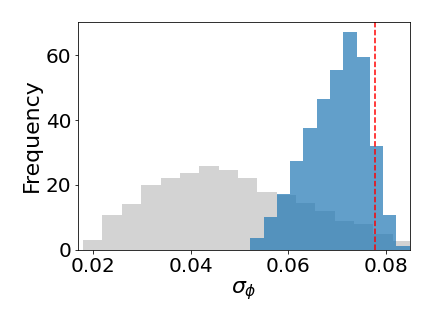}}\\
\subfloat[$E_s$ (both data types)]{\label{E_s}\includegraphics[width = 50mm]{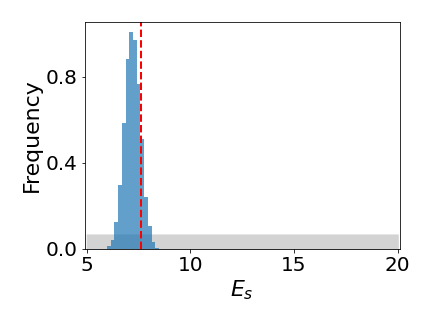}}
\hspace{10mm}
\subfloat[$E_o$ (both data types)]{\label{E_o}\includegraphics[width = 50mm]{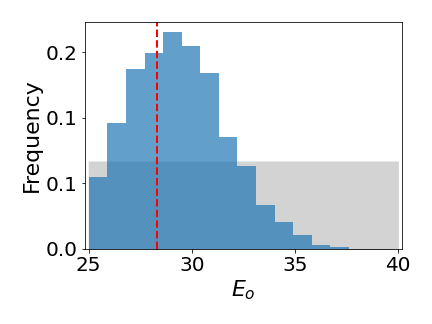}}
\caption{History matching results for true model~2 using the total error covariance $C_{\mathrm{tot}}$ in the likelihood function (Eq.~24 in the main paper). Gray regions represent prior distributions, blue histograms are posterior distributions using both data types, and red vertical lines denote true values. Legend in (a) applies to all subplots.}
\label{meta_true_2}
\end{figure}

\begin{figure}[!ht]
\centering   
\subfloat[True model~2]{\label{logk_True_2}\includegraphics[width=31mm]{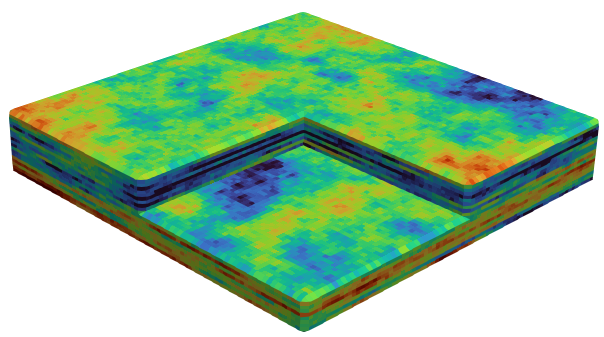}}
\hspace{1mm}
\subfloat[Posterior 1]{\label{Posterior_logk_1_True_2}\includegraphics[width=31mm]{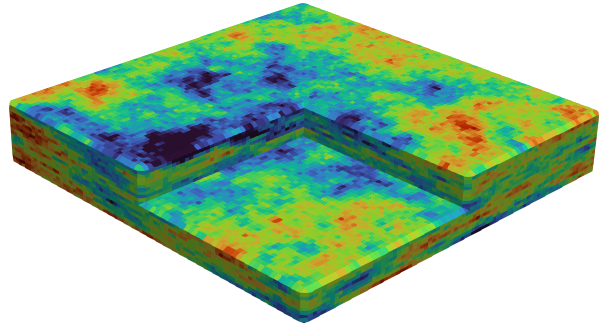}}
\hspace{1mm}
\subfloat[Posterior 2]{\label{Posterior_logk_2_True_2}\includegraphics[width=31mm]{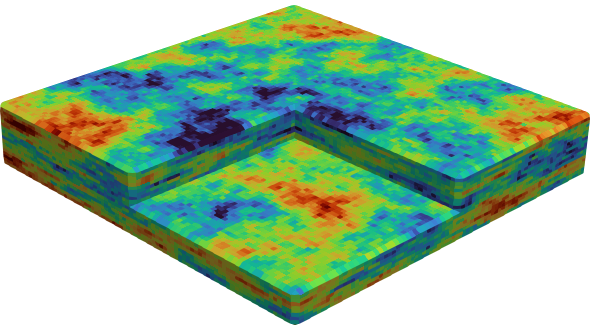}}
\hspace{1mm}
\subfloat[Posterior 3]{\label{Posterior_logk_3_True_2}\includegraphics[width=31mm]{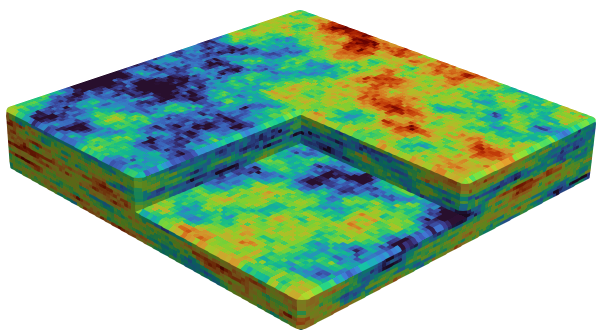}}
\hspace{1mm}
\subfloat[Posterior 4]{\label{Posterior_logk_4_True_2}\includegraphics[width=31mm]{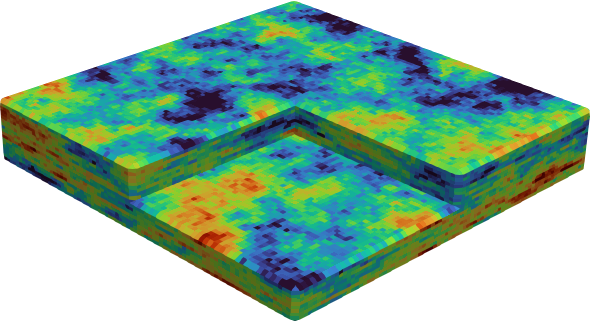}}
\includegraphics[width=5.5mm]{Figure/Surrogate/logk_Scale.PNG}
\caption{Representative log-permeability fields for posterior geomodels. Log-permeability field for true model~2 shown in (a).}
\label{Prior_Posterior:logk_True_2}
\end{figure}

The posterior results for the metaparameters for this case are shown in Fig.~\ref{meta_true_2}. Significant uncertainty reduction is achieved, and the true values are within the posterior distributions despite the true model falling outside the prior in terms of correlation structure. Two of the quantities that are predicted less accurately in Fig.~\ref{meta_true_2} are $\sigma_{\log k}$ and $\log_{10} a_r$. For a given value of $\log_{10} a_r$, the lower vertical correlation length in true model~2 could tend to reduce vertical communication relative to prior models, thus causing the MCMC procedure to underestimate $\log_{10} a_r$ (which is what we observe in Fig.~\ref{meta_true_2}c). The lower $l_v$ could also lead to more variability, causing an overestimation of $\sigma_{\log k}$ (as observed in Fig.~\ref{meta_true_2}b). 

The true model, along with four representative posterior log-permeability fields of the storage aquifer, are shown in Fig.~\ref{Prior_Posterior:logk_True_2}. Although there is general resemblance in terms of property range between the true model and these posterior realizations, we do observe a higher degree of horizontal correlation and a lower degree of variability in Fig.~\ref{Prior_Posterior:logk_True_2}a relative to Fig.~\ref{Prior_Posterior:logk_True_2}b-e. The latter effect is consistent with the posterior results for $\sigma_{\log k}$ in Fig.~\ref{meta_true_2}b.

\begin{figure}[!ht]
\centering  
\subfloat[True]{\label{s_true_2}\includegraphics[width = 30mm]{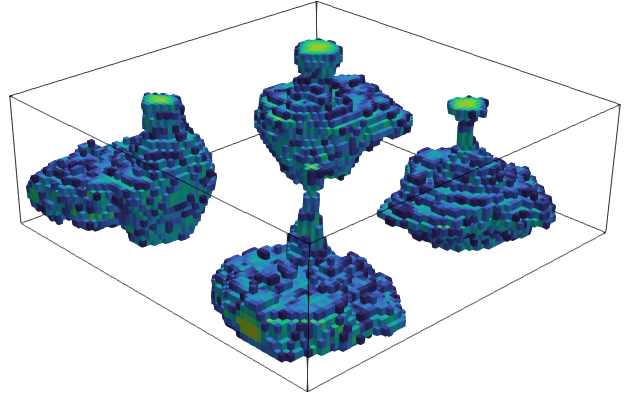}}
\hspace{1.5mm}
\subfloat[Posterior 1]{\label{posterior_s_1_1}\includegraphics[width=30mm]{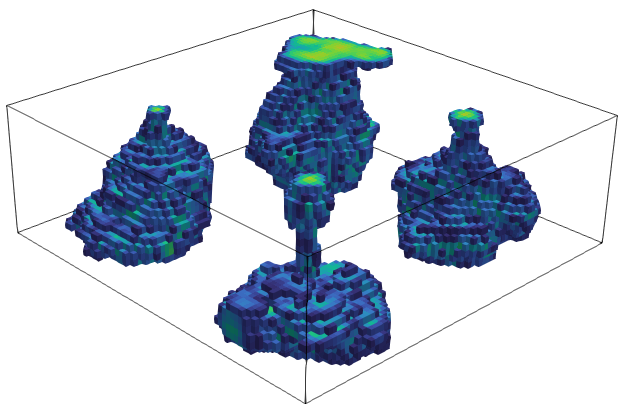}}
\hspace{1.5mm}
\subfloat[Posterior 2]{\label{posterior_s_1_2}\includegraphics[width=30mm]{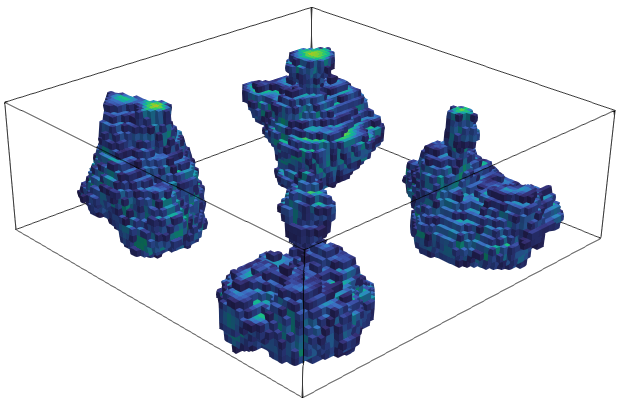}}
\hspace{1.5mm}
\subfloat[Posterior 3]{\label{posterior_s_1_3}\includegraphics[width=30mm]{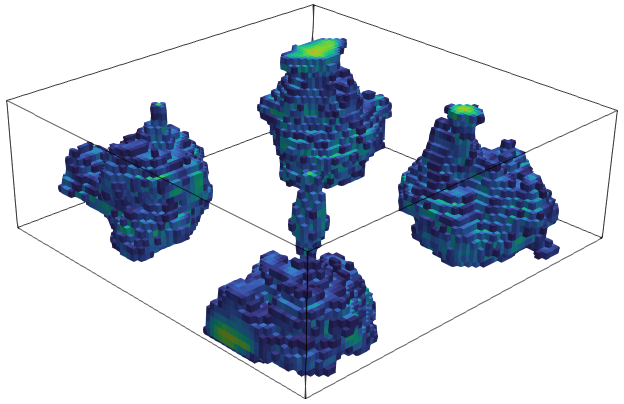}}
\hspace{1.5mm}
\subfloat[Posterior 4]{\label{posterior_s_1_4}\includegraphics[width=30mm]{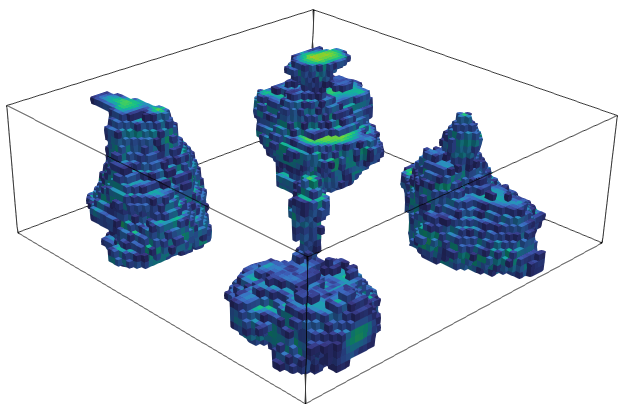}}
\hspace{1.5mm}
\includegraphics[width=6.5mm]{Figure/Surrogate/Sw_Scale.PNG}
\caption{Representative CO$_2$ saturation fields at 30~years from posterior geomodels (b--e)  for true model~2 using both data types. True saturation field at 30~years shown in (a).}
\label{Prior_Posterior_1:Saturation}
\end{figure}

\begin{figure}[!ht]
\centering   
\subfloat[True]{\label{d_true_2}\includegraphics[width=32mm]{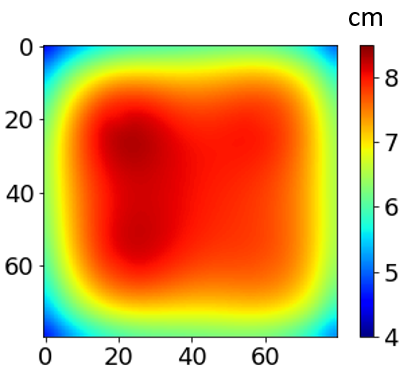}}
\hspace{2mm}
\subfloat[Posterior 1]{\label{posterior_d_1_1}\includegraphics[width=32mm]{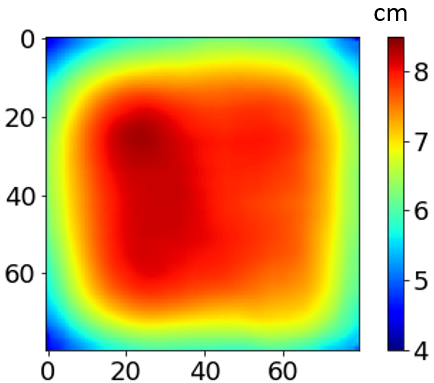}}
\hspace{2mm}
\subfloat[Posterior 2]{\label{posterior_d_1_2}\includegraphics[width=32mm]{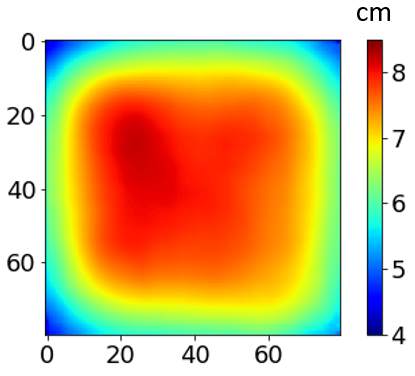}}
\hspace{2mm}
\subfloat[Posterior 3]{\label{posterior_d_1_3}\includegraphics[width=32mm]{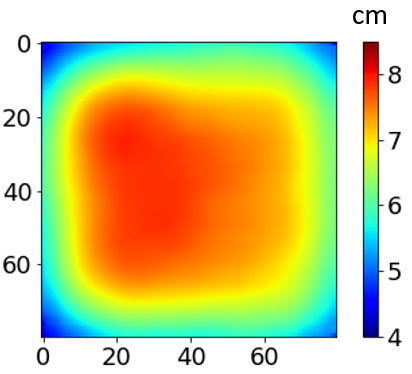}}
\hspace{2mm}
\subfloat[Posterior 4]{\label{posterior_d_1_4}\includegraphics[width=32mm]{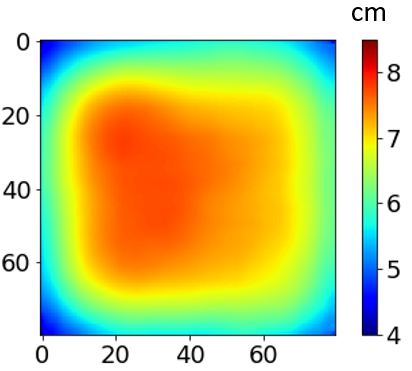}}
\caption{Representative surface displacement fields at 30~years from posterior geomodels (b--e) for true model~2 using both data types. True surface displacement field at 30~years shown in (a).}
\label{Prior_Posterior_1:Displacement}
\end{figure}

Four representative posterior saturation and surface displacement fields at 30~years are shown in Figs.~\ref{Prior_Posterior_1:Saturation} and~\ref{Prior_Posterior_1:Displacement}. The true results are also shown in these figures. The prior saturation fields (upper row in Fig.~17 in the main paper) and surface displacement fields (upper row in Fig.~19 in the main paper) are the same as for true model~1. The true plumes in Fig.~\ref{Prior_Posterior_1:Saturation}a are very irregular and quite different from the plumes for true model~1 (Fig.~9a in the main paper). The posterior fields capture many of the key saturation features, including the thin upper portion of the front plume. The posterior surface displacement fields show quite close agreement with the true solution in Fig.~\ref{d_true_2}. The overall amount of displacement and the slightly larger uplift in the left portion of the model are captured in the four posterior results in Fig.~\ref{Prior_Posterior_1:Displacement}.

The results for true model~2 are quite acceptable even though the model is outside the prior. This suggests the general hierarchical MCMC procedure possesses a degree of robustness. If  data are not well predicted by the calibrated models, the workflow could be extended to include modification of the prior metaparameters or their distributions, and/or the update of model error (covariance) as applied by \citet{oliver2018calibration}. Modification of the prior would require some amount of surrogate model retraining, but our framework is compatible with an iterative workflow of this type.

\bibliographystyle{Main} 
\bibliography{ref}

\end{document}